\newif\ifvldb
\ifvldb
\documentclass{vldb}
\else
\documentclass{vldb-mod}
\fi

\usepackage{graphicx}
\usepackage{balance}  

\usepackage[utf8]{inputenc} 
\usepackage[T1]{fontenc}    
\usepackage{url}            
\usepackage{booktabs}       
\usepackage{amsfonts}       
\usepackage{nicefrac}       
\usepackage{microtype}      
\usepackage{bm}
\usepackage{graphicx}
\usepackage{subfigure}
\usepackage{algorithm}
\usepackage{xspace}
\usepackage{color,soul}
\usepackage{multirow}
\usepackage{multicol}
\usepackage[labelfont=bf]{caption}

\newcommand{\specialcell}[2][c]{\begin{tabular}[#1]{@{}c@{}}#2\end{tabular}}

\usepackage{enumitem}
\setenumerate[1]{itemsep=0pt,partopsep=0pt,parsep=\parskip,topsep=0pt,leftmargin=10pt}
\setitemize[1]{itemsep=0pt,partopsep=0pt,parsep=\parskip,topsep=0pt,leftmargin=10pt}

\DeclareMathOperator*{\argmax}{arg\,max}

\newcommand{\alg}{\text{Vero}\xspace}

\vldbTitle{An Experimental Evaluation of Large Scale GBDT Systems}
\vldbAuthors{Fangcheng Fu, Jiawei Jiang, Yingxia Shao, Bin Cui}
\vldbDOI{https://doi.org/10.14778/3342263.3342273}
\vldbVolume{12}
\vldbNumber{11}
\vldbYear{2019}

\begin{document}


\title{An Experimental Evaluation of Large Scale GBDT Systems}



%
%
%
%

\numberofauthors{1} 

\author{
%
%
\alignauthor
Fangcheng Fu{$^{\S\#}$}~~~Jiawei Jiang{$^{\S\#}$}~~~Yingxia Shao{$^{\dag}$}~~~Bin Cui{$^{\S\P}$}
\\\affaddr{
{$^\S$}Department of Computer Science and Technology \& Key Laboratory of High Confidence Software Technologies (MOE), Peking University\\
{$^\P$}Center for Data Science \& National Engineering Laboratory for Big Data Analysis and Applications, Peking University\\
{$^\dag$}Beijing University of Posts and Telecommunications
~~~{$^\#$}Tencent Inc.
}
\\\email{
\{ccchengff,bin.cui\}@pku.edu.cn~~~shaoyx@bupt.edu.cn~~~jeremyjiang@tencent.com}
}

\maketitle

\begin{abstract}
Gradient boosting decision tree (GBDT) is a widely-used machine learning algorithm 
in both data analytic competitions and real-world industrial applications.
Further, driven by the rapid increase in data volume,
efforts have been made to train GBDT in a distributed setting
to support large-scale workloads.
However, we find it surprising that
the existing systems manage the training dataset in different ways, but 
none of them have studied the impact of data management.
To that end, this paper aims to study the pros and cons of different data management methods 
regarding the performance of distributed GBDT.

We first introduce a quadrant categorization of data management policies
based on data partitioning and data storage.
Then we conduct an in-depth systematic analysis  
and summarize the advantageous scenarios of the quadrants.
Based on the analysis, we further propose a novel distributed GBDT system named {\alg},
which adopts the unexplored composition of vertical partitioning and row-store 
and suits for many large-scale cases. 
To validate our analysis empirically, we implement different quadrants in the same code base 
and compare them under extensive workloads,
and finally compare {\alg} with other state-of-the-art systems over a wide range of datasets.
Our theoretical and experimental results provide a guideline on 
choosing a proper data management policy for a given workload.
\end{abstract}

\section{Introduction}
\label{sec:intro}

Gradient boosting decision tree (GBDT)~\cite{friedman2001greedy}
is an ensemble model which uses decision tree as weak learner and improves model quality with a boosting strategy~\cite{friedman2000additive,Wang2012}.
It has achieved superior performance in various workloads, such as
prediction, regression, and ranking~\cite{li2012robust,tyree2011parallel,burges2010ranknet}. Not only the data scientists 
choose it as a favorite tool for data analytic 
competitions such as Kaggle, but also users from industry raise interests in deploying
GBDT in production environments~\cite{he2014practical,zhou2017psmart,jiang2018dimboost}.

With the rapid increase in data volume,
distributed GBDT has been intensively studied to improve the performance. Recently, 
a range of distributed machine learning systems has been developed to train GBDT, such as XGBoost, LightGBM and DimBoost~\cite{chen2016xgboost,TencentBoost,zhou2017psmart,ponomareva2017tf,ke2017lightgbm,jiang2018dimboost}.
However, in practical use, there is no such system 
able to outperform the others in all cases.
We notice that these systems manage the training dataset in different ways.
This motivates us to conduct a study of the data management in distributed GBDT.

Consider the training dataset as a matrix, where each row represents one instance
and each column refers to one dimension of feature.
To make distributed machine learning possible, we need to partition the dataset
among the workers in a cluster.
Afterwards, each worker uses some storage structure to store the data partition.
As a result, there are two orthogonal aspects in the data management of distributed GBDT --- data partitioning and data storage.

\textbf{Data Partitioning.}
Since the dataset is a two-dimensional matrix,
there are two different schemes to partition the dataset over the workers.
\textbf{Horizontal partitioning}, which is the de facto choice of most distributed machine learning algorithms, horizontally partitions the dataset by instances (rows). 
\textbf{Vertical partitioning} is an alternative to horizontal partitioning.
The workers partition the dataset by features (columns)
and each worker stores a feature subset.

\textbf{Data Storage.}
After data partitioning, each worker has a portion of the training data, either a horizontal partition or a vertical partition.
Without loss of generality, we assume the dataset is sparse.
There are two avenues to store the data.
\textbf{Row-store} is a popular choice in machine learning.
Each instance is stored as a set of 
$\langle$feature index, feature value$\rangle$pairs, a.k.a. Compressed Sparse Row (CSR) format.
Many algorithms follow a row-based training routine which supports scanning the training data sequentially.
\textbf{Column-store} puts together one column (feature) of the partition, and stores each column as a set of $\langle$instance index, feature value$\rangle$ pairs, a.k.a. Compressed Sparse Column (CSC) format.

\begin{figure}[!t]
  \centering
  \includegraphics[width=2in]{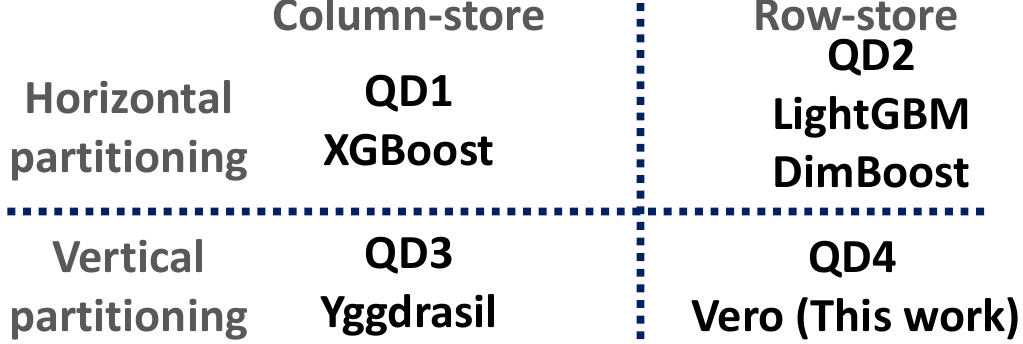}
  \caption{Quadrants of existing works}
  \label{fig:four_quadrants}
\end{figure}

If we revisit the methods of data management, there are two data partitioning choices and
two data storage choices, yielding 
four possible combinations.
Using a quadrant-based manner, Figure~\ref{fig:four_quadrants} summarizes four combinations into four quadrants.
Interestingly, three quadrants have been explored by existing systems,
but none of these works study which is the best combination.
As a result, the researchers and engineers might be confused 
when they need to choose the platform for their specific workloads.
To address this issue, we ask the question
\textit{what are the advantages and disadvantages of different data management schemes, and how can we make a proper choice facing different scenarios?}

\subsection{Summary of Contributions}
We list the main contributions of this work below.

\textbf{\underline{(Anatomy of existing systems)}}
To answer the above questions, we first study how data management influences the performance of distributed GBDT.
Specifically, we conduct a theoretical analysis of data partitioning and data storage.

\textit{Anatomy of data partitioning.}
The data partitioning directly affects the communication and memory cost 
due to a data structure called gradient histogram, 
which summarizes gradient statistics for fast and accurate split finding in GBDT.
We find that vertical partitioning is more suitable for a range of workloads, 
including high-dimensional features, deep trees, and multi-classification.
The fundamental reason is that these factors could cause extreme large gradient histograms,
and vertical partitioning helps avoid intensive communication and memory overhead.
In contrast, horizontal partitioning works better for datasets with low dimensionality and a large number of instances.

\textit{Anatomy of data storage.}
In GBDT, the training procedures, especially the construction of gradient histograms involve complex data access and indexing, and the efficiency is influenced by the data storage.
We carefully investigate the computation efficiency of row- and column-store
in terms of data access and indexing.
We find that although column-store seems more natural for
vertical partitioning, as adopted by database design, the computation overhead is rather undesirable.
Row-store is superior to column-store
given a large number of instances, achieving a higher computation efficiency.
In short, our main finding is that row-store is almost always a wiser choice 
unless the dataset is high-dimensional and meanwhile contains very few instances.

\textbf{\underline{(Proposal of {\alg})}}
Unfortunately, although our study discovers that the fourth quadrant in Figure{~\ref{fig:four_quadrants}}
is suitable for a wide range of large-scale scenarios, 
including high-dimensional datasets, multi-classification tasks, and deep trees, 
it is never investigated by previous works.
In this work, we propose {\alg}, an end-to-end distributed GBDT system that 
uses vertical partitioning and row-store.

\textit{Horizontal-to-vertical transformation.}
We develop an efficient algorithm to transform the 
horizontally stored datasets to vertically partitioned.
To reduce the network overhead, we compress both feature indices and feature values, 
without any loss of model accuracy.

\textit{Training with Vertical Row-store.}
We redesign the training routine of GBDT to match the vertical partitioning and row-store policy.
Specifically, we adapt the split finding and node splitting procedures to vertical partitioning, 
and adopts a node-to-instance index for row-store to construct the gradient histograms efficiently.

\textbf{\underline{(Comprehensive Evaluation)}}
We implement distributed GBDT on top of Spark{~\cite{zaharia2012resilient}}, 
a popular distributed engine for large-scale data processing,
and conduct extensive experiments to validate our analysis empirically.

\textit{Breakdown comparison of data management.}
To fairly evaluate each candidate in data partitioning and data storage,
we implement different partitioning schemes and storage patterns in the same code base, 
and compare them under different circumstances using a wide range of datasets.
Our experimental results regarding computation, communication, and memory cost validate our theoretical anatomy.

\textit{End-to-end evaluation.}
We compare {\alg} with other popular GBDT systems over extensive datasets, 
including public, synthetic, and industrial datasets.
Empirical results show that our analytical comparison also holds for 
the state-of-the-art systems.
Regarding the results, we provide suggestions on how to choose 
a proper platform for a given workload.

\section{Background}
\label{sec:bg}

\subsection{Preliminaries of GBDT}
\subsubsection{Overview of GBDT}
\begin{figure}[!t]
  \centering
  \includegraphics[width=2.3in]{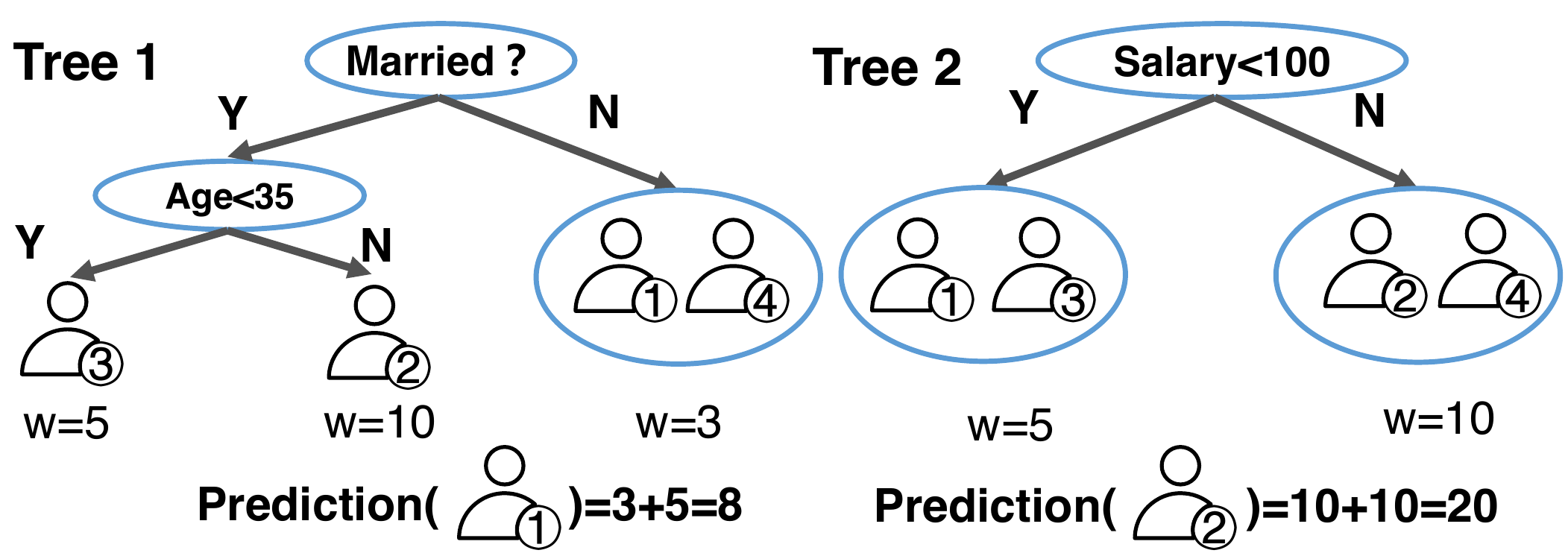}
  \caption{An illustration of GBDT}
  \label{fig:gbdt}
\end{figure}

Gradient boosting decision tree is a boosting algorithm that uses decision tree as weak learner.
Figure~\ref{fig:gbdt} shows an illustration of GBDT.
Given a training dataset with $N$ instances and $D$ features
$\left\{(\bm{x}_i, y_i)\right\}_{i=1}^N$, 
where $\bm{x}_i \in \mathbb{R}^{D}$ and $y_i \in \mathbb{R}$ 
are the feature vector and label of an instance, 
GBDT trains a set of decision trees
$\left\{f_t(\bm{x})\right\}_{t=1}^{T}$, 
puts each instance onto one leaf node,
and sums the leaf predictions of all trees as the final instance prediction:
$\hat{y}_i = \sum_{t=1}^T \eta f_t(\bm{x}_i)$, 
where $T$ denotes the total number of trees and $\eta$ is a hyper-parameter
called learning rate (a.k.a. step size).

GBDT trains the decision trees sequentially. For the $t$-th tree,
it tries to minimize the loss given the predictions of prior trees, 
defined by the regularized objective function:

\begin{equation*}
\small
    F^{(t)} \text{=} \sum l(y_i, \hat{y}_i^{(t)}) + \Phi(f_t)
    \text{=} \sum l(y_i, \hat{y}_i^{(t-1)} + f_t(\bm{x}_i)) + \Phi(f_t),
\end{equation*}

where $l$ is usually a differentiable convex loss function that measures 
the loss given prediction and target, e.g., logistic loss or square loss.
$\Phi$ is a regularization term to avoid over-fitting. We follow 
the popular choice in~\cite{chen2016xgboost,jiang2018dimboost}, which is
$\Phi(f_t) = \gamma J_t + 
\lambda||\omega_t||_2^2 / 2$,
where $\omega_t$ denotes the weight vector comprised of 
$J_t$ leaf values in in the $t$-th tree.
$\gamma$ and $\lambda$ are hyper-parameters that control the complexity
of one tree.

To quickly optimize the objective function, LogitBoost~\cite{friedman2000additive}
proposes to approximate $F^{(t)}$ with second-order 
Taylor expansion when training the $t$-th tree, i.e., 

\begin{equation*}
\small
    F^{(t)} \approx \sum \big[ l(y_i, \hat{y}_i^{(t-1)}) + g_i f_t(\bm{x}_i) + \frac{1}{2}h_i f_t^2(\bm{x}_i) \big] + \Phi(f_t),
\end{equation*}
where $g_i=\partial_{\hat{y}_i^{(t-1)}} l(y_i, \hat{y}_i^{(t-1)})$ 
and $h_i=\partial_{\hat{y}_i^{(t-1)}}^2 l(y_i, \hat{y}_i^{(t-1)})$
are the first- and second-order gradients. 
Denote $I_j=\{i|\bm{x}_i \in {leaf}_j\}$ as the set of instances 
classified onto the $j$-th leaf. Omitting the constant term, we should minimize

\begin{equation*}
\small
    \widetilde{F}^{(t)} =\sum_{j=1}^{J_t}\Big[ \big(\sum_{i\in{I_j}}g_i\big) \omega_j 
    + \big(\sum_{i\in{I_j}}h_i + \lambda\big) \omega_j^2 \Big] + \gamma J_t.
\end{equation*}

If the tree is not going to be expanded (no leaf to be split), 
we can obtain its optimal weight vector and minimal loss by

\begin{equation}
\small
\label{eq:criterion}
    \omega_j^*=-\frac{\sum_{i\in{I_j}}g_i}{\sum_{i\in{I_j}}h_i+\lambda},
    \widetilde{F}^{(t)*}=-\frac{1}{2}\sum_{j=1}^{J_t}\frac{(\sum_{i\in{I_j}}g_i)^2}{\sum_{i\in{I_j}}h_i+\lambda}+\gamma J_t.
\end{equation}

Equation~\ref{eq:criterion} can be reckoned as a measurement to evaluate
the performance of a decision tree, which can be analogous to
the impurity functions of decision tree algorithms, such as 
entropy for ID3~\cite{quinlan1986induction} or 
Gini-index for CART~\cite{breiman2017classification}.
To grow a tree w.r.t. minimizing the total loss, 
the common approach is
to select a tree node (beginning with the root node) and 
find the best split (a split feature and a split value) that can achieve the maximal split gain.
The split gain is defined as

\begin{equation}
\small
\label{eq:split_gain}
    Gain=\frac{1}{2}\big[ \frac{(\sum_{i\in{I_L}}g_i)^2}{\sum_{i\in{I_L}}h_i+\lambda} + \frac{(\sum_{i\in{I_R}}g_i)^2}{\sum_{i\in{I_R}}h_i+\lambda} - \frac{(\sum_{i\in{I}}g_i)^2}{\sum_{i\in{I}}h_i+\lambda} \big] - \gamma,
\end{equation}
where $I_L$ and $I_R$ indicate the left and right child nodes after the splitting.
After the current tree finishes, the predictions of all instances are updated, the gradient statistics are re-computed, and the algorithm will proceed to next tree.

\begin{figure}[!t]
  \centering
  \includegraphics[width=3.3in]{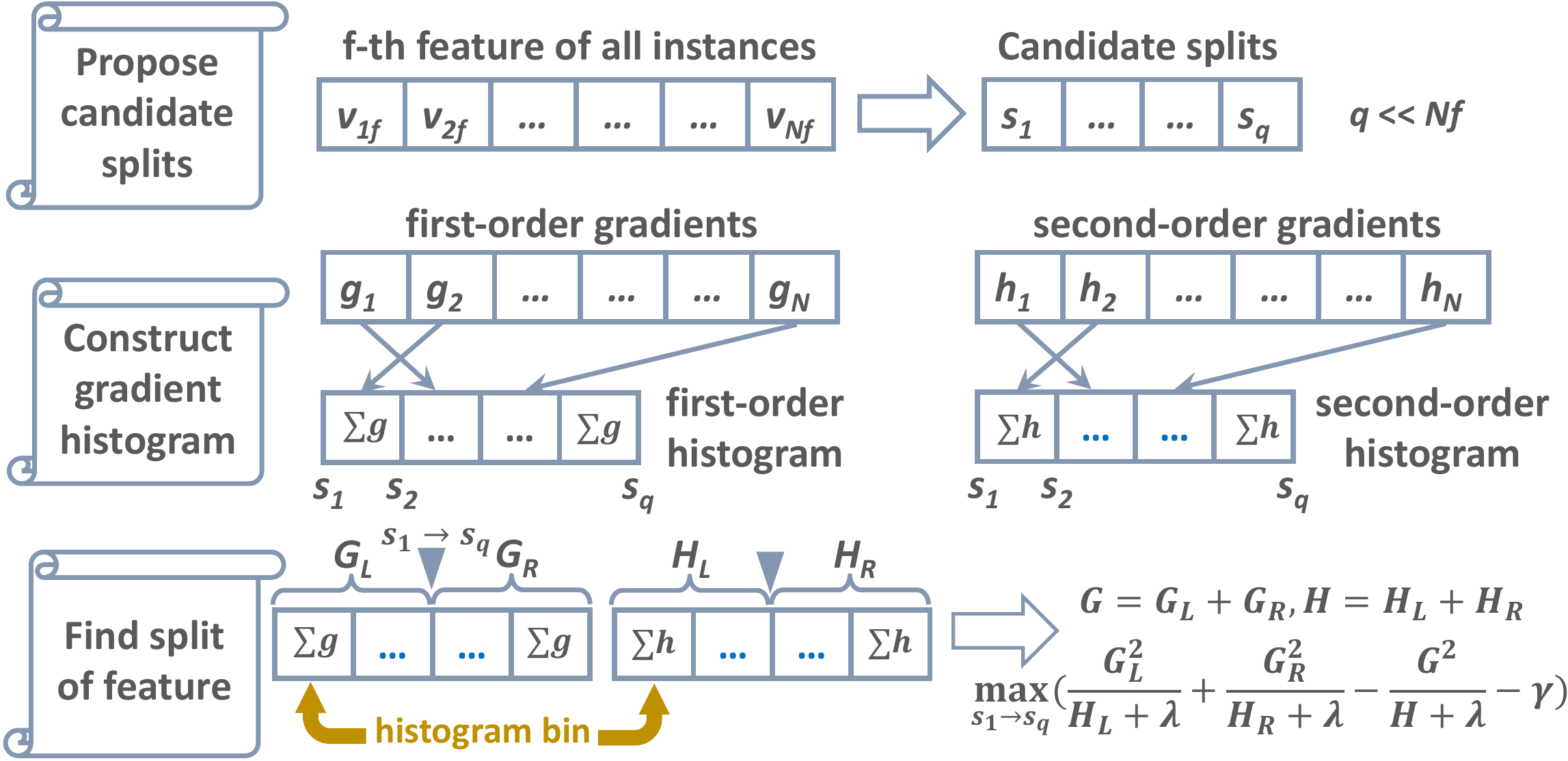}
  \caption{Histogram-based split finding for one feature}
  \label{fig:histogram}
\end{figure}

\subsubsection{Histogram-based Algorithm}

\textbf{Histogram-based split finding.}
It is vital to find the optimal split of a tree node efficiently, 
as enumerating every possible split in a brute-force manner is impractical. 
Current works generally adopt a histogram-based algorithm for fast and accurate split finding, as illustrated in Figure{~\ref{fig:histogram}}.
The algorithm considers only $q$ values for each feature $f$ as candidate splits rather than all possible splits.
The most common approach to propose the candidates is using the 
quantile sketch~\cite{greenwald2001space,karnin2016optimal,gan2018moment}
to approximate the feature distribution.
After candidate splits are prepared, we enumerate all instances on a tree node and accumulate their gradient statistics into two histograms, first- and second-order gradients, respectively.
The histogram consists of $q$ bins, each of which sums the first- or second-order gradients
of instances whose $f$-th feature values fall into that range.
In this way, each feature is summarized by two histograms.
We find the best split of 
feature $f$ upon the histograms by Equation~\ref{eq:split_gain}, and the global best split
is the best split over all features.

\textbf{Histogram subtraction technique.}
Another advantage of the histogram-based algorithm is that we can accelerate 
the algorithm by a histogram subtraction technique.
The instances on two children nodes are non-overlapping and mutual exclusive,
since an instance will be classified onto either left or right child node when the parent node gets split.
Considering the basic operation of histogram is adding gradients, therefore, 
for feature $f$, the element-wise sum of first- or second-order histograms of children nodes
equals to that of parent. 
Motivated by this, we can significantly accelerate training by 
first constructing the histograms of the one child node with fewer instances, 
and then getting those of the sibling node via histogram subtraction (histograms of parent node are persist in memory).
By doing so, we can skip at least one half of the instances.
Since histogram construction  usually dominates the computation cost,
such subtraction technique can speed up the training process considerably.

\begin{figure}[!t]
	\centering
    \subfigure[{Horizontal partitioning. Workers construct local histograms for all features and aggregate into global ones.}]{
    \label{fig:horizontal}
    \includegraphics[width=220pt]{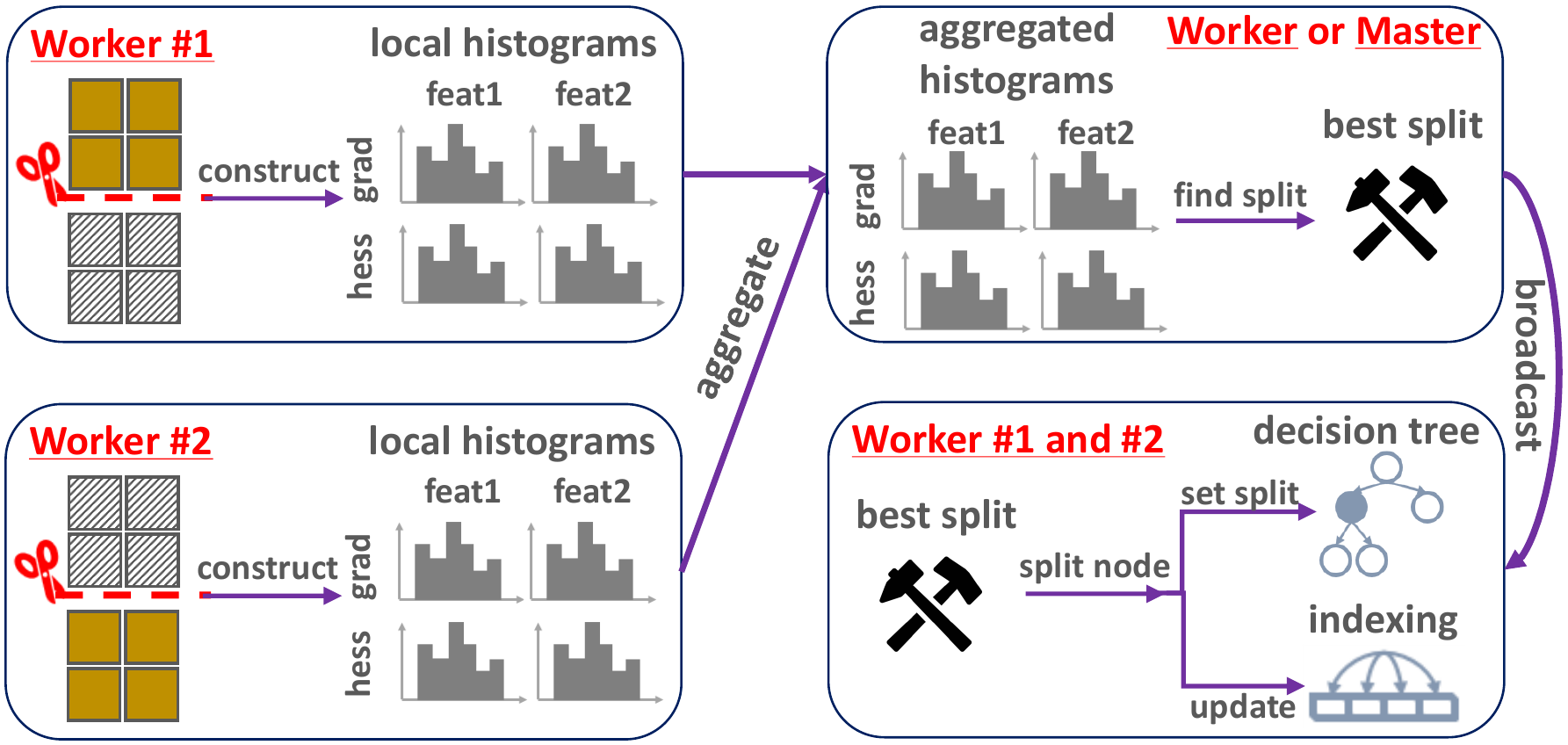}}
    \subfigure[{Vertical partitioning. Worker who proposes global best split broadcasts the instance placement after node splitting.}]{
    \label{fig:vertical}
    \includegraphics[width=240pt]{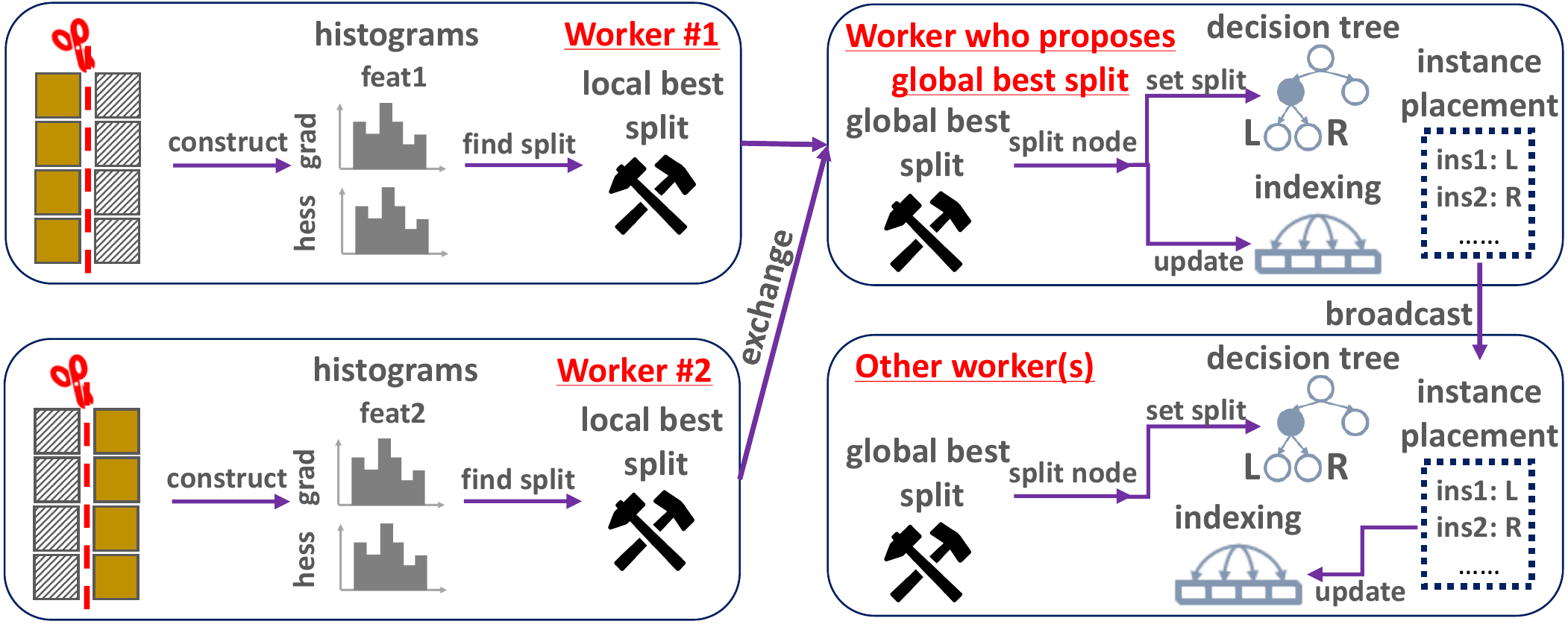}}
    \subfigure[{Row-store and column-store}]{
    \label{fig:storage}
    \includegraphics[width=2.6in]{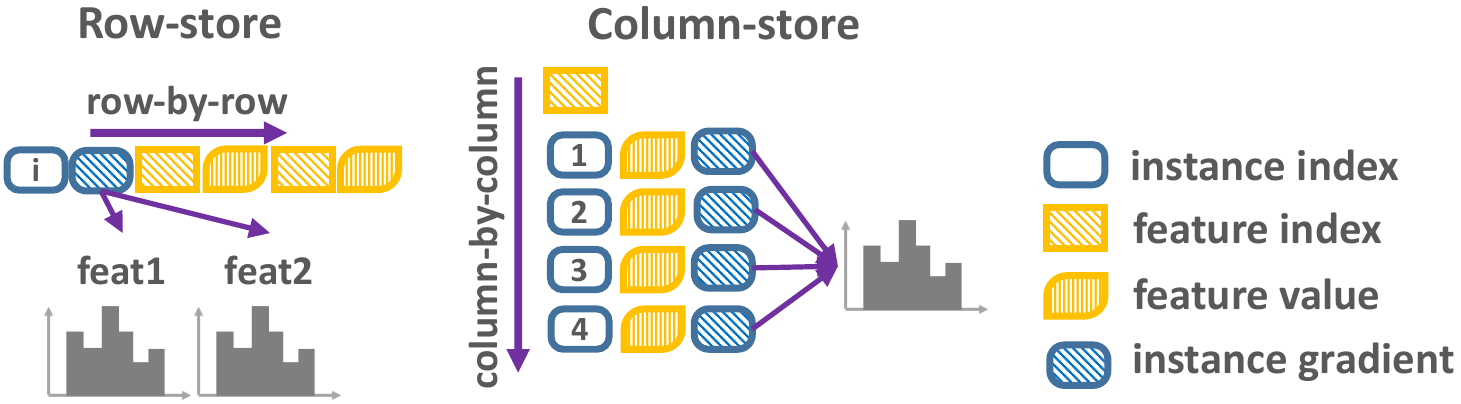}}
	\caption{{Illustration of data partitioning and storage}}
	\label{fig:diff}
\end{figure}

\subsection{Data Management in GBDT}
As aforementioned, the combinations of partitioning schemes and storage patterns together form four quadrants (QD).
Although the four quadrants entail similar memory consumption to store the dataset in expectation,
the manipulation (including computation, storing, and communication) of gradient histograms 
can be significantly different.

\subsubsection{Data Partitioning in GBDT}
Since gradient histograms can be reckoned as summaries of features, 
different partitioning choices affect the way we construct and exchange histograms.

Since values of each feature are scattered among workers in horizontal partitioning,
as presented in Figure~\ref{fig:horizontal}, 
each worker needs to construct histograms for all features 
based on its data shard.
Then the local histograms are aggregated into global histograms via element-wise summation,
so that all values of each feature are correctly summarized.

As shown in Figure~\ref{fig:vertical}, 
each worker maintains one or several complete columns in vertical partitioning, 
therefore there is no need to aggregate the histograms.
Each worker obtains the local best split regarding its feature subset, 
and then all workers exchange the local best splits and choose the global best one.
Nevertheless, since the feature values of an instance are partitioned, 
its placement after node splitting, 
i.e., left or right child node, 
is only known by the worker who proposes the global best split.
As a result, the placement of instances must be broadcast to all workers.

\subsubsection{Data Storage in GBDT}
The most distinguished difference brought by storage pattern is the way 
we index and access the values during the construction of histograms, 
as shown in Figure{~\ref{fig:storage}}.

With row-store, each worker iterates the data shard row-by-row, 
and accumulates the gradient statistics to corresponding histograms.
When processing one instance, the worker needs to update multiple histograms of different features.
To accelerate the construction,
each worker further maintains an indexing between tree nodes and instances.

With column-store, as all values of one feature are held together, 
each worker constructs histograms one-by-one by processing the columns individually.
However, the indexing between the values on a column 
and tree nodes must be maintained carefully.
As we will discuss in Section{~\ref{sec:storage_pattern}}, 
the data access and indexing might take extra efforts.

\section{Anatomy of Quadrants}
\label{sec:anatomy}

In this section, we provide an in-depth study of the four quadrants when
training a GBDT model distributedly.
To formally describe the results, we assume there are $W$ workers,
and the GBDT model is comprised of $T$ decision trees, where each of them has $L$ layers.
The number of candidate splits is denoted by $q$.
For classification tasks, we denote $C$ as the dimension of 
a gradient, where $C$ equals 1 in binary-classification 
or the number of classes in multi-classification.

\subsection{Analysis of Partitioning Scheme}
Here we theoretically analyze the performance of
horizontal and vertical partitioning schemes, including memory and communication cost. 

\subsubsection{Histogram Size}
\label{sec:hist_size}
The core operation of GBDT is the construction and manipulation of gradient histograms.
We first study the size of histograms, which is determined by three factors.
(1) {\bf Feature dimension.} Since two histograms are built for 
    each feature (one first-order gradient histogram and one second-order gradient histogram), the total size is proportional to $2 \times D$.
(2) {\bf Number of candidate splits.} The number of bins in one histogram 
    equals to the number of candidate splits $q$, 
    which makes the histogram size proportional to $q$.
(3) {\bf Number of classes.}  In multi-classification tasks, 
    the gradient is a vector of partial derivatives on all classes.
    The histogram size is therefore proportional to $C$.
To sum up, the histogram size on one tree node, denoted by $Size_{hist}$, is 
$2 \times D \times q \times C \times 8$ bytes, where $8$ bytes is 
the size of a double-precision floating-point number.

\subsubsection{Memory Cost}
Obviously, the memory cost for both partitioning 
to store the dataset is similar. 
Nonetheless, the memory cost to store the gradient histograms is quite different.
Here we focus on the memory consumed by storing the histograms.

In order to perform histogram subtraction, we have to conserve
the histograms of the parent nodes.
The maximum number of histograms to be held in memory 
equals to the number of tree nodes in the last but one layer~\footnote{We assume all histograms are preserved in memory.},
which is $2^{L-2}$.
With horizontal partitioning, each worker needs to
construct the histograms of all features, thus the
memory cost of histograms is $Size_{hist} \times 2^{L-2}$. 
Nevertheless, with vertical partitioning, each worker
constructs the histograms of a portion of features.
As a result, the expected memory cost is
$Size_{hist} \times 2^{L-2}/{W}$, which is
significantly smaller than the horizontal partitioning counterpart.

\subsubsection{Communication Cost}
The dominant communication cost in horizontal partitioning scheme
is the aggregation of histograms.
Despite the existence of different aggregation methods~\cite{thakur2005optimization}, such as
\texttt{map-reduce}, \texttt{all-reduce}, and \texttt{reduce-scatter},
the minimal transferred data of each worker is the size of
local histograms. 
Thus the total communication cost among the cluster building one tree is at least
$Size_{hist} \times W \times (2^{L-1} - 1)$. 
It is obvious that as the tree goes deeper, i.e., as $L$ increases, the communication cost grows quadratically.

Unlike horizontal partitioning scheme, vertical partitioning scheme does not 
need to aggregate the histograms since each worker
holds all the values of a specific feature.
However, as described in Section~\ref{sec:bg}, after splitting a tree node,
the placement of instances must be broadcast to all workers. 
Since the communication cost is
only affected by the number of instances,
the overhead in one tree layer remains the same as the tree goes deeper. 
As we will elaborate in Section~\ref{sec:train_workflow}, the placement is encoded into a bitmap so that 
the communication overhead can be reduced sharply.
To conclude, the communication cost for an $L$-layer tree is
$\left\lceil {N}/{8} \right\rceil \times W \times L$ bytes,
where $\left\lceil {N}/{8} \right\rceil$ bytes is the size of one bitmap.

\subsubsection{Summary of Analysis}

Undoubtedly, the choice of partitioning scheme highly depends on $Size_{hist}$.
Undoubtedly, horizontal partitioning works well for datasets with low dimensionality,
since the resulting histograms are small.
However, in both industry and academia, the following three cases become more and more popular --- high dimensional features, deep trees, and multi-classification.
In these cases, the histogram size can be very large.
Therefore, vertical partitioning is far more memory- and communication-efficient than
horizontal partitioning.
Take an industrial dataset \textit{Age} as an example, which is also used in our 
experimental study, we suppose running GBDT on 8 workers.
The dataset contains 48M instances, 330K features and 9 classes.
The decision trees have 8 layers and the number of candidate splits is 20.
Then the estimated size of histograms on one tree node can be up to 906MB.
Using the horizontal approach, the memory consumption would be 56.6GB and the total
communication cost would be 900GB for merely one tree in the worst case.
To the contrary, when the vertical scheme is applied, the expected memory cost
of histograms is 7.08GB per tree and the communication cost is merely 366MB for one tree.

\subsection{Analysis of Storage Pattern}
\label{sec:storage_pattern}
In this section, we discuss the impact brought by different storage patterns. 
Although there exist various works discussing the different 
storage patterns in database designs, the conclusion cannot be transferred to distributed GBDT.

The choice of storage pattern only influences
the computation cost, rather than communication or memory cost.
The most time-consuming computation in GBDT is histogram construction.
However, the data access in GBDT is different from other ML models.
Specifically, since GBDT conducts tree splitting in a top-to-bottom way, 
we need to create an index between tree nodes and training instances, and
update the index during the training.
Below, we discuss how to design the index with different storage patterns.

\begin{figure}[!t]
  \centering
  \includegraphics[width=3in]{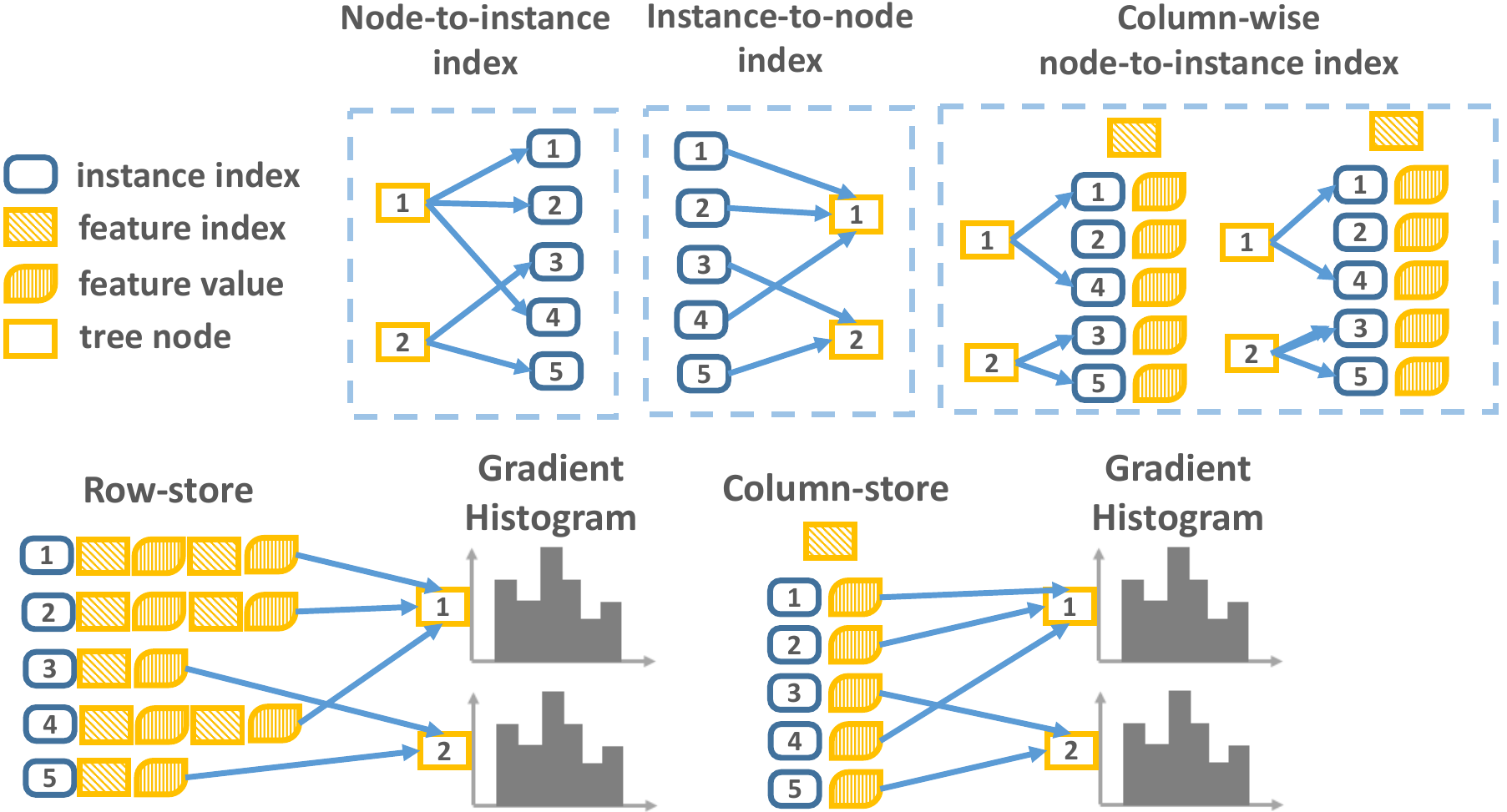}
  \caption{Illustration of different indexes}
  \label{fig:indexes}
\end{figure}

\subsubsection{Choice of Index}

To understand the computation complexity of histogram construction,
we first illustrate the possible index choices used in GBDT training.
As illustrated in Figure~\ref{fig:indexes}, there are three commonly used indexes indicating the position of training instances in the tree.

\begin{itemize}
    
    \item {\bf Node-to-instance index}
    maps a tree node to the corresponding training instances, meaning that
    the key is a tree node and the value is the instances on the tree node.

    \item {\bf Instance-to-node index}
    maps a training instance to the corresponding tree node.
    
    \item {\bf Column-wise node-to-instance index}
    maintains a node-to-instance index for each feature column.
    
\end{itemize}

\subsubsection{Row-store}

When building the gradient histogram with row-store, we adopt a row-wise access method to scan rows sequentially.
Each row is an instance, which consists of the instance index and a list of
nonzero $\langle$feature id, feature value$\rangle$ pairs.

\textbf{Node-to-instance index} is designed for row-store.
We get the instance rows of one tree node from the index.
For each row, we iterate the $\langle$feature id, feature value$\rangle$ pairs.
For each pair, we add the instance gradient to the histograms of that tree node.
Furthermore, the node-to-instance index enables the histogram subtraction technique since we can directly get the instances of any tree node. 
If two tree nodes are siblings, we only build histogram for the tree node with fewer instances,
and apply histogram subtraction for the other one.
Consequently, combining the node-to-instance index and row-store can save large amount of data accesses.

\subsubsection{Column-store}

When building the gradient histogram with column-store, a straight-forward way is to use a column-wise access method to scan the columns.
Each column summarizes the values of one feature, which includes the feature id and a list of $\langle$instance id, feature value$\rangle$ pairs.

\textbf{Instance-to-node index.}
Since the key of each pair in column-store is instance id, 
a natural idea is creating an instance-to-node index.
As shown in Figure~\ref{fig:indexes}, for each $\langle$instance id, feature value$\rangle$ pair,
we query the tree node it belongs to, and then update the corresponding histograms.
Nonetheless, we find that using such method is not efficient in practice.
The reason is that in many real cases, the dataset is often sparse 
(especially for high-dimensional datasets).
By default, given an optimal node split with feature $f$, instances with missing value on $f$ are classified to the same child node, causing imbalance sibling nodes.
Histogram subtraction should be able to boost the performance, however, 
with instance-to-node index, we cannot directly 
get the instances of two child nodes without queries, 
i.e., we need to access all instances of the two nodes.
Therefore, a lot of time is wasted on scanning unnecessary data, resulting in
poor performance.

\textbf{Node-to-instance index.}
One solution to avoid scanning all instances is using node-to-instance index for column-store. 
However, there still exists a fatal drawback.
Once obtaining an instance id from the index,
we need to locate the feature values of the instance from column-store.
To that end, we have to perform a binary search 
on all the feature columns, which brings in a $\log{(N)}$ computation complexity.
When $N$ is large, the overhead becomes unacceptable.

\begin{figure}[!t]
  \centering  
  \includegraphics[width=1.7in]{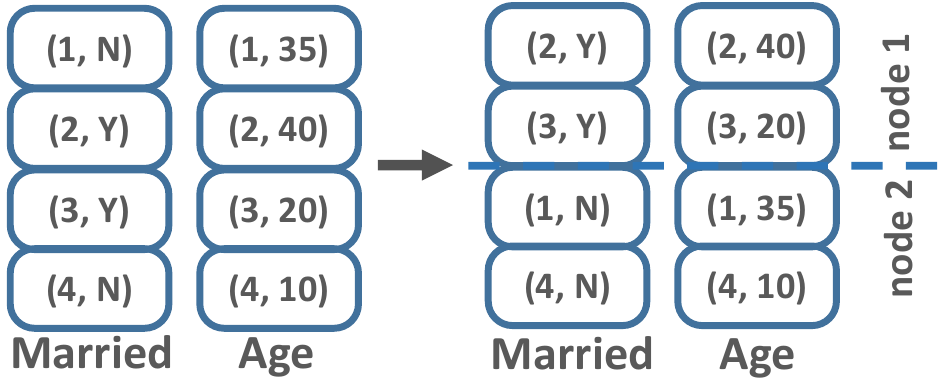}
  \caption{Update of column-wise node-to-instance index (w.r.t. the first tree in Figure~\ref{fig:gbdt}).}
  \label{fig:update_index}
\end{figure}

\textbf{Column-wise node-to-instance index.}
Another way to escape from both scanning unnecessary data and binary search is 
deploying an index on each column, 
which actually maintains a node-to-instance index for each column.
When building histograms for one node, we can locate the $\langle$instance id, feature value$\rangle$ pairs on all columns directly.
Nevertheless, although locating the instances is fast, updating the index is expensive.
As shown in Figure~\ref{fig:update_index},
whenever we split some tree node, we have to update the indexes on all columns.
The computation complexity of splitting tree nodes is about $D$ times of the two indexes described above.
As a result, the column-wise node-to-instance index is only applicable for low-dimensional datasets.

\begin{table*}[!tb]
\scriptsize
\centering
\caption{Summary of advantageous scenarios among different quadrants.}
\begin{tabular}{c c c c c c c c c}
\toprule[1.5pt]
\multirow{2}*{Quadrants} & \multicolumn{2}{c}{Technique} & \multicolumn{4}{c}{Data Characteristics} & \multicolumn{2}{c}{Model} \\
\cmidrule[0.5pt](l{1pt}r{1pt}){2-9}
& Partitioning & Storage & High dim. & Low dim. & High ins. & Low ins. & Multi-class & Deep tree \\
\midrule[1pt]
QD1 & Horizontal & Column & & & & & & \\
\midrule[1pt]
QD2 & Horizontal & Row & & \checkmark & \checkmark & & & \\
\midrule[1pt]
QD3 & Vertical & Column & \checkmark & & & \checkmark & & \\
\midrule[1pt]
QD4 & Vertical & Row & \checkmark & & \checkmark & & \checkmark & \checkmark \\
\bottomrule[1.5pt]
\end{tabular}
\label{tb:adv_summary}
\end{table*}

\subsubsection{Summary of Analysis}

Here we summarize the computation complexity of different combinations 
by considering the number of accesses to dataset or other data structures.

{\bf Cost of histogram construction.}
In histogram construction, since we need to access the feature values on the data shard, 
and the expected number of key-value pairs is $Nd/W$, 
where $d$ is the average number of non-zeros of one instance,
the complexity of histogram construction for one layer is at least $O(Nd/W)$.
There are three combinations that can theoretically achieve the lowest complexity, 
which are row-store with node-to-instance index, column-store with instance-to-node index, 
and column-store with column-wise index.
However, as discussed above, column-store with instance-to-node index 
cannot benefit from the histogram subtraction technique, and thereby 
spends more time than row-store with node-to-instance index in practice;
while column-store with column-wise index 
entails a much higher complexity when node splitting
although it works well for histogram construction.
For the last combination, column-store with node-to-instance index, 
it incurs binary search on the feature columns whenever accessing an instance.
In expectation, the complexity of binary search is approximately $O(\log Nd/WD)$.
Therefore, the overall complexity becomes $O(Nd/W\times\log Nd/WD)$.

{\bf Cost of split finding and node splitting.}
Except for histogram construction, there are two other phases in GBDT, which are split finding and node splitting. 
To make the analysis self-contained, here we briefly analyze the computation cost in these two phases.
For split finding, the algorithm needs to iterate all split candidates, causing a computation complexity of $O(qD/W)$, regardless of the partitioning scheme.
For node splitting, we need to update the index described above.
The computation on one tree layer for both store patterns is proportional to the number of instances, if we do not use the column-wise node-to-instance index~\footnote{The complexity of column-wise node-to-instance index is $O(Nd/W)$, so we exclude it from our consideration.}.
The complexity is $O(N/W)$ for horizontal partitioning and $O(N)$ for vertical partitioning.
Obviously, both of the two phases have a significantly lower computation cost than histogram construction.
Therefore, we should pay more attention to the impact of storage pattern on histogram construction.

{\bf Summary.}
As analyzed, column-store is not
efficient with different index structures.
To the contrary, the combination of row-store and node-to-instance index
can achieve minimal computation since it 
leverages histogram subtraction to reduce instance scanning 
and incurs the smallest cost of index update.
As a result, unless the dataset contains very few instances
so that the extra cost in indexing will not be large,
we should choose row-store for distributed GBDT.

\subsection{Take-away Results}
We conclude the advantageous scenarios of different data management methods in Table{~\ref{tb:adv_summary}}.
Considering large-scale cases is becoming more and more ubiquitous, 
we have the following take-away results:

\begin{itemize}
    \item 
    Vertical partitioning is able to outperform horizontal partitioning 
    for the high-dimensional features, deep trees and multi-classification tasks,
    since it is more memory- and communication-efficient, 
    while horizontal partitioning is better the low-dimensional datasets.

    \item
    Row-store is better than column-store unless the number of instances is very small, 
    since it can achieve minimal computation complexity and avoid redundant data accesses.
    
    \item
    Overall, the composition of vertical partitioning and row-store (QD4) 
    achieves optimal performance
    under many real-world large-scale cases as aforementioned.
    In Section{~\ref{sec:eval}} and{~\ref{sec:industrial}},
    we will validate this through extensive experiments.
\end{itemize}

\section{Representatives of Quadrants}

In this section, we first introduce the representatives of QD1-3, 
and then propose {\alg}, a brand new distributed GBDT system with vertical partitioning and row-store (QD4).

\subsection{Taxonomy of Existing Systems}
\label{sec:existing}

\textbf{XGBoost (QD1, Horizontal \& Column)}.
XGBoost~\cite{chen2016xgboost} is a popular GBDT system that 
achieves great success, and it chooses
horizontal partitioning scheme and column-store pattern. 
In XGBoost, each worker maintains an instance-to-node indexing.
To construct histograms of one layer, workers linearly scan the feature columns,
accumulate the gradient statistics to corresponding histogram bins,
and finally aggregate the histograms in an \texttt{all-reduce} manner.
After aggregation, the histograms are owned by a leader worker.
Then it finds the best split by enumerating the candidate splits in the histograms.
In node splitting phase, each worker updates its own instance-to-node index.

\textbf{LightGBM and DimBoost (QD2, Horizontal \& Row)}.
Both LightGBM~\cite{ke2017lightgbm} and DimBoost~\cite{jiang2018dimboost} 
belong to this quadrant. A node-to-instance indexing that 
maps tree nodes to instances is maintained.
To construct the histograms of one node, the workers scan the feature vectors
of instances on that node, accumulate the gradient statistics to corresponding histogram bins,
and finally aggregate the histograms.
LightGBM accomplishes the aggregation using \texttt{reduce-scatter}.
Instead of aggregating all histograms on a single worker,
each worker is responsible for a part of features.
All workers then find splits on aggregated histograms 
and synchronize to obtain the global best one.
While DimBoost, with \texttt{parameter-server} architecture~\cite{li2014scaling,jiang2017heterogeneity}, aggregates the histograms
on parameter servers and enables server-side split finding.
In either way we can avoid the single-point-bottleneck in communication. 
The node-to-instance indexing is also updated during node splitting.

\textbf{Yggdrasil (QD3, Vertical \& Column)}.
Although Yggdrasil~\cite{abuzaid2016yggdrasil} is designed for
vanilla decision tree algorithms instead of GBDT, it is the first work that introduces
vertical partitioning into distributed decision tree.
In Yggdrasil, each worker maintains several complete columns so that 
it can obtain the best split of its own feature (column) subset without histogram aggregation.
All workers then exchange their local best splits and choose the
global best with maximal split gain.
In this way, the communication in split finding phase is far less than
horizontal-based methods.
When splitting the tree nodes, Yggdrasil encodes the placement of each instance into a bitmap.
Further, Yggdrasil utilizes a column-wise node-to-instance index.
Based on the bitmap, the index for each column is updated.
However, it will bring in a large computation cost when feature dimensionality is high.

\begin{figure}[!t]
  \centering
  \includegraphics[width=3in]{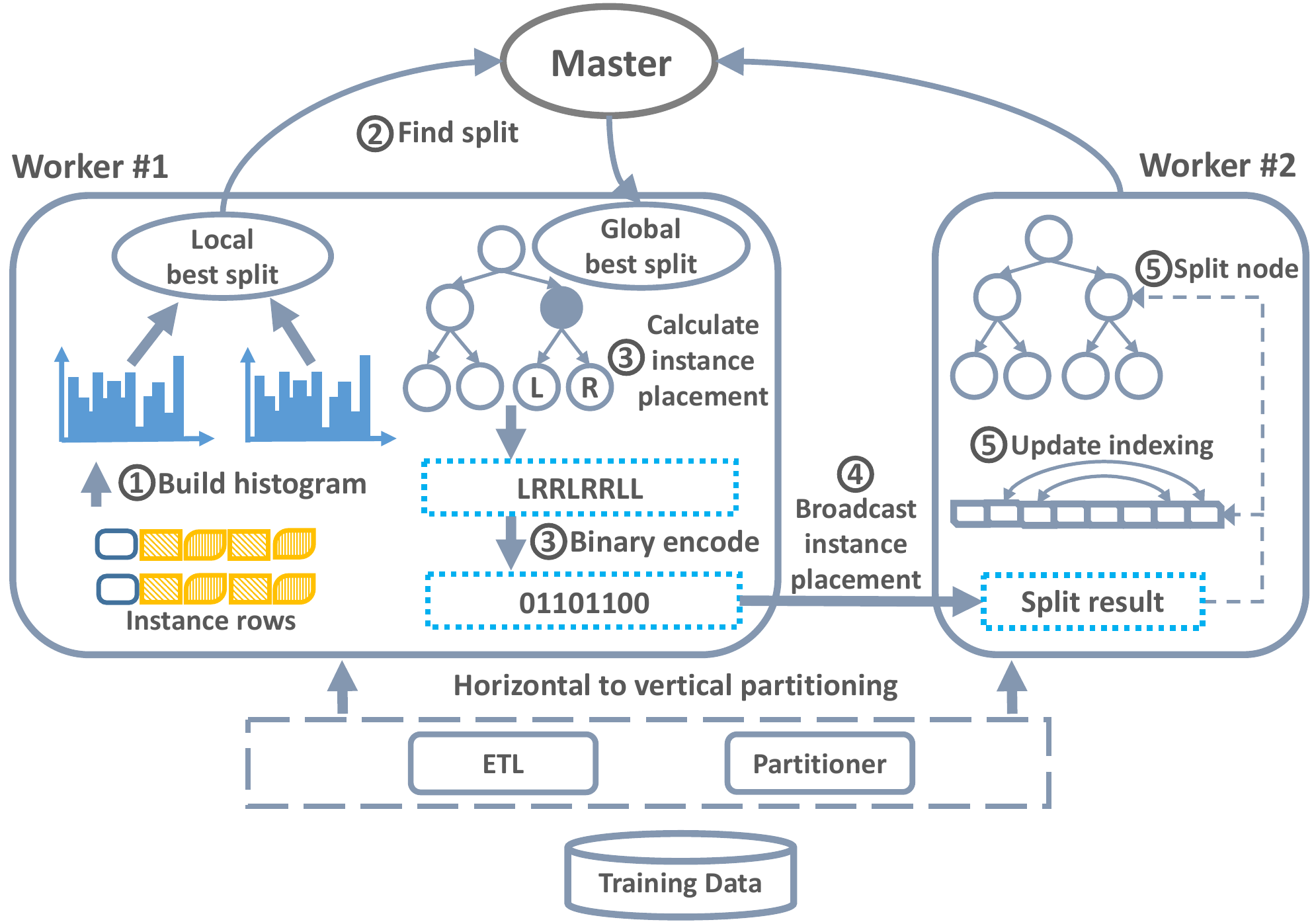}
  \caption{Overview of \alg}
  \label{fig:overview}
\end{figure}

\subsection{\alg}
As analyzed in Section{~\ref{sec:anatomy}}, 
\textbf{QD4 (Vertical \& Row)} is superior to the others 
under many large-scale scenarios but left unexplored.
This drives us to develop a system, {\alg}, within the scope of QD4.
{\alg} is built on top of Spark~\cite{zaharia2012resilient} 
and has been deployed in our industrial partner, Tencent Inc..
As shown in Figure~\ref{fig:overview},
{\alg} follows the master-worker architecture.
After loading horizontally partitioned dataset from distributed file systems, 
we perform an efficient transformation operation to vertically repartition the dataset accross workers.
Then masters and workers iteratively train a set of decision trees upon the repartitioned dataset.

\subsubsection{Horizontal-to-Vertical Transformation}
\label{sec:transform}
Naturally, training datasets are often horizontally partitioned and
stored in distributed file systems such as HDFS and S3, which is
obviously unfit for vertical partitioning. To solve this problem,
we need to repartition the datasets vertically.
To address the potential network overhead for large datasets, 
we develop an efficient transformation method that compresses both 
feature indices and feature values, without any loss of model accuracy.
There are five main steps, as shown in Figure~\ref{fig:transform} and described below.

\begin{enumerate}[wide,leftmargin=0pt,labelindent=0pt]

\item\textbf{Build quantile sketches.}
After loading the dataset, each worker builds a quantile sketch for each feature.
Then the local sketches are repartitioned among all workers, i.e., 
the local sketches of one feature are sent to the same worker.
Finally, the workers merge local sketches of 
the same feature into a global sketch.

\item\textbf{Generate candidate splits.}
The workers generate candidate splits for each feature from merged quantile sketch, using a set of quantiles, e.g., {0.1, 0.2, ..., 1.0}.
Then the master collects the candidate splits and broadcasts them
to all workers for further use.

\begin{figure}[!t]
  \centering
  \includegraphics[width=3in]{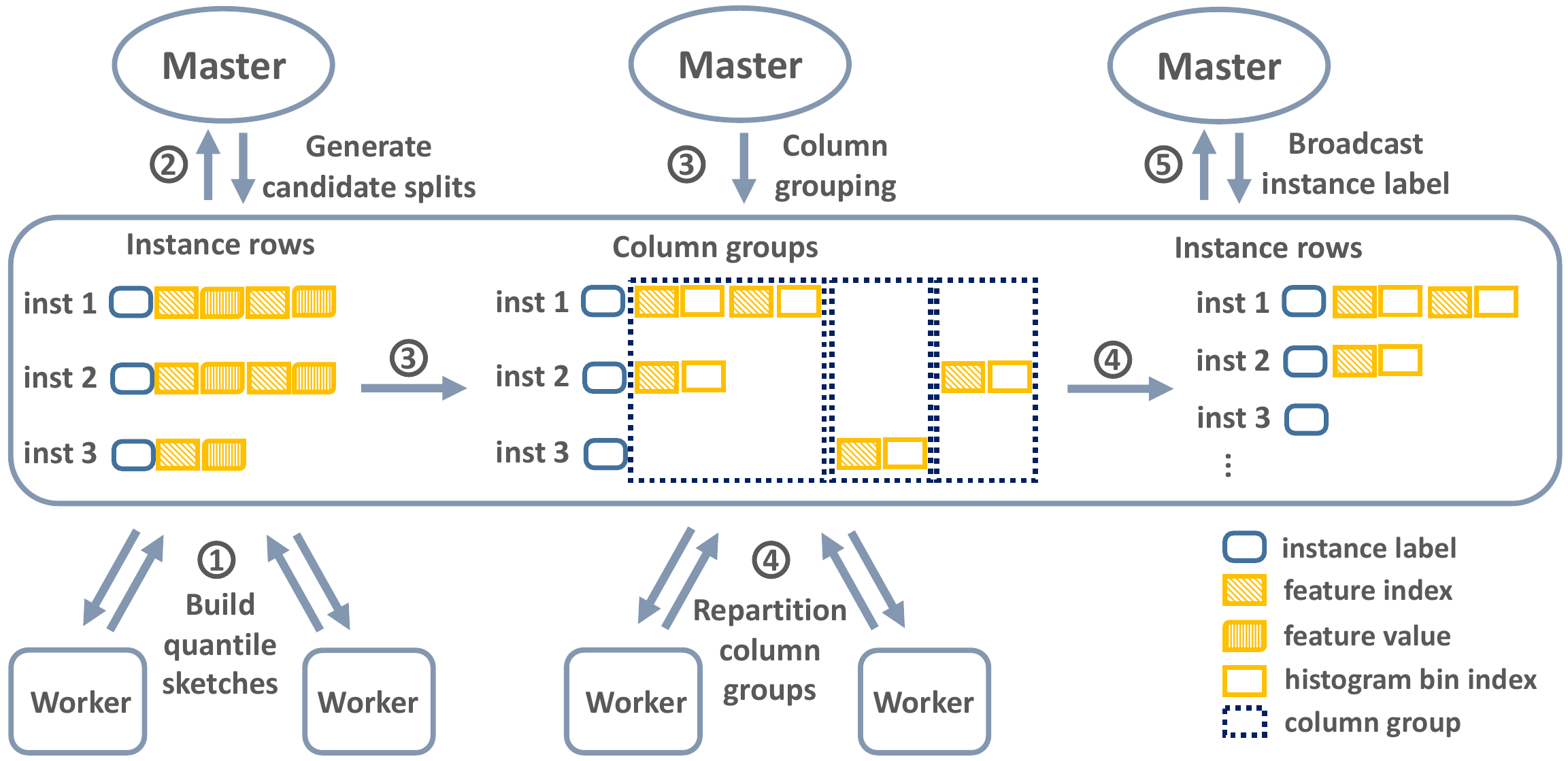}
  \caption{Horizontal to vertical transformation}
  \label{fig:transform}
\end{figure}

\item\textbf{Column grouping.}
Each worker changes the representation of its local data shard
by putting those features to be assigned to the same worker into one group.
(The strategy of feature assignment will be described in Section~\ref{sec:transform_impl}.)
The key-value pairs are encoded into a more compact form simultaneously.
(i) For each feature, we assign a new feature id starting from 0 inside the column group.
Suppose there are $p$ features in one group, we use $\lceil \log (p) \rceil$ bytes to encode the new feature id.
(ii) We encode feature values with histogram bin indexes, 
which indicates the range of two consecutive splits.
Since the histograms stay unchanged, the model accuracy will not be harmed.
As the number of histogram bins $q$ is generally a small integer, 
we further encode bin indexes with $\lceil \log (q) \rceil$ bytes.
After this operation, key-value pairs turn into $\langle$new feature id, bin index$\rangle$ pairs.

\item\textbf{Repartition column groups.}
Similar to step 1, the column groups are repartitioned among workers.
By doing so, each worker holds all values of its responsible features.
Further, the ordering of instances should be the same on all workers, so that we can coalesce the instances with their labels.
This can be done by sorting the received column groups w.r.t. the original worker ids.

\item\textbf{Broadcast instance labels.}
Master collects all instance labels and
broadcasts them to all workers.
Since the instance rows on each worker are ordered in step 4, we can therefore coalesce instance rows with instance labels. 

\end{enumerate}

\textbf{Network overhead.}
Step 1 and 2 prepare the candidate splits for step 3 to convert
feature values into bin indexes. 
Quantile sketch is a widely-used data structure for approximate query~\cite{Li2018,Song2018} and is usually small in size~\cite{greenwald2001space,karnin2016optimal,gan2018moment}, 
so the network overhead is almost negligible.
The communication bottleneck incurs in step 4.
Nevertheless, by encoding feature id and feature value into smaller bytes,
the size of a key-value pair is significantly decreased.
According to our empirical results, it brings up to 4$\times$ compression.
The time cost of step 5 is not dominant as presented in 
the appendix of our technical report~\cite{1907.01882}.

\subsubsection{Training Workflow}
\label{sec:train_workflow}
To fit the data management strategy of QD4, we revise the traditional training procedure of GBDT.

\textbf{Histogram construction.}
Given tree node(s) to process,
the master first obtains the number of instances on each node, 
then it decides on which node(s) we can perform histogram subtraction and sends the schema to all workers.
Each worker constructs histograms based on its data shard.
Since \alg stores data in row manner, we use the node-to-instance index to achieve the best performance in histogram construction.
For each tree node, each worker obtains a list of row indexes, and each row represents an instance that is currently classified onto that tree node.
Then the worker adds the gradient statistics to 
corresponding histograms.
We also adopt the method proposed in~\cite{jiang2018dimboost} 
to handle instances with missing values.
Finally, unlike horizontal-based works, \alg does not need to aggregate histograms among workers.

\textbf{Split finding.}
To obtain the best split for some tree node,
each worker first calculates a split for each histogram by Equation~\ref{eq:split_gain},
and proposes the one with maximal split gain as the local best split.
Finally, master collects all local best splits and chooses the global best one.
Note that, the obtained feature id is not the original one
since we transform it in step 3 of Section~\ref{sec:transform}.
Hence, the master needs to recover the original feature afterwards.

\textbf{Node splitting.}
As aforementioned, since only one worker owns feature values of the best split,
the placement of each instance (left or right child) after node splitting can only be computed by it.
The master asks the worker who has proposed the global best split
to compute and broadcast the instance placement.
Since the placement of each instance has only two options, i.e., left or right child node, 
we use a bitmap to represent the instance placement,
which can reduce the network overhead by 32$\times$.
All workers then update the node-to-instance index based on the bitmap.

\subsubsection{Proposed Optimization}
\label{sec:transform_impl}

\textbf{Load balance.}
There are various strategies for column grouping, such as round robin, hash-based, 
and range-bashed partition, yet these methods cannot guarantee exact load balance.
We might suffer from the straggler problem if a worker contains far more key-value pairs than others.
Therefore, we balance the workload on workers by averaging the total number of key-value pairs. 
In practice, the master collects the number of feature occurrences 
from global quantile sketches,
then the problem becomes assigning the feature pairs to $W$ groups so that the number of feature pairs in each group is as close as possible.
This problem is obviously an NP-hard problem, we therefore use a greedy method to solve it~\cite{jiang2013predictive}.

\begin{figure}[!t]
  \centering
  \includegraphics[width=3.5in]{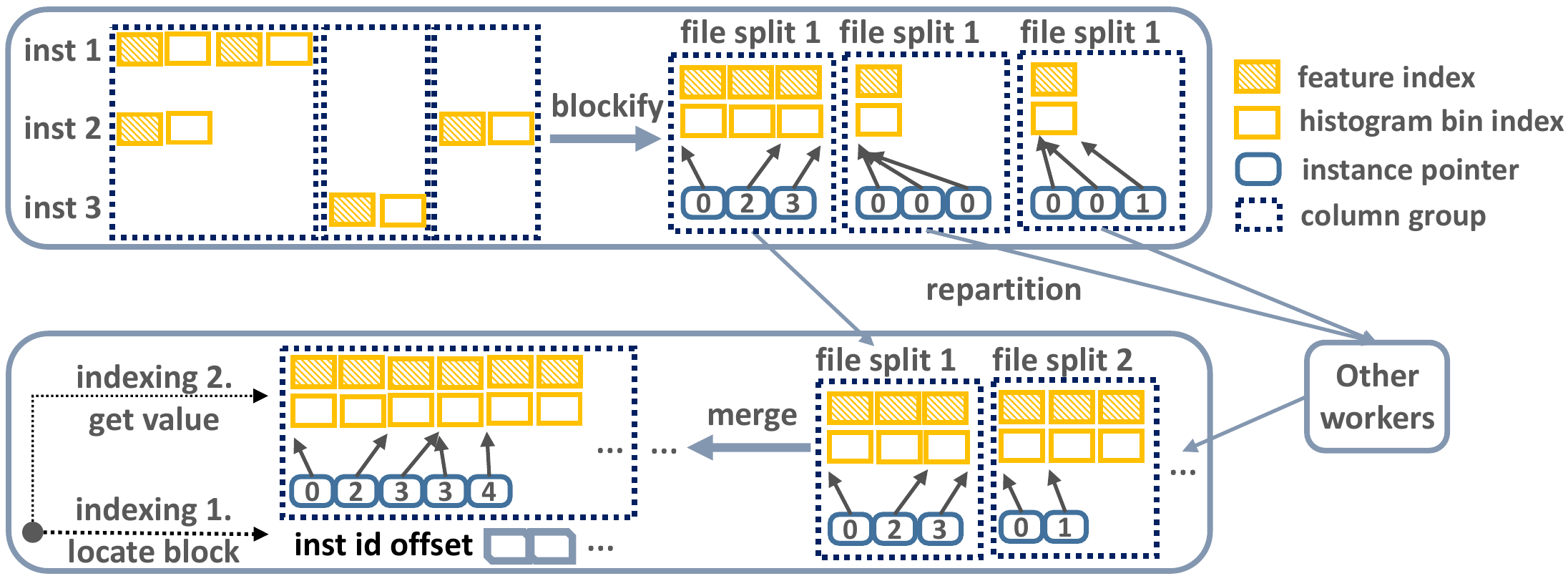}
  \caption{\small Blockfied column grouping and two-phase indexing}
  \label{fig:blocky}
\end{figure}

\textbf{Blockify of column group.}
Although the network overhead is reduced by compression,
the overhead of (de)serialization is probably large
if we represent column groups with large amount of small vectors, 
since there are $W$ times number of objects compared to the original dataset.
To alleviate such overhead, we blockify 
the column groups before repartition,
as shown in Figure~\ref{fig:blocky}.
Each block consists of three arrays, i.e.,
feature indexes, histogram bin indexes, and instance pointers.
By default, the file split in Spark is 128MB, therefore,
we can always put a partial column group into one block 
since the number of key-value pairs in one file split is 
far smaller than \texttt{INT\_MAX}.
We assign the index of file split to the $W$ partial column groups.
After repartition, each column group (the data sub-matrix of a worker)
is comprised of several blocks, sorted by their file split indexes.

\textbf{Two-phase indexing and block merge.}
Since the data sub-matrix is now made up of a number of blocks,
we adopt a two-phase index to access each instance.
In initialization, the offset of instance (row) id of each block is recorded.
Given an instance id, we first binary search the block
that contains that instance, then the instance id
inside the block is calculated by subtracting the offset of the block, finally we obtain the range of the instance
by the instance pointers.
Considering that the number of file splits can be very large,
for instance, a 100GB dataset results in approximately 800 file splits,
we merge the blocks when possible in order to reduce the 
data access time.
In practice, the number of blocks after the merge operation is smaller than 5.
Therefore, we can nearly omit the extra cost brought by two-phase indexing.

\section{Evaluation}
\label{sec:eval}

In this section, we conduct experiments to empirically validate our analysis.
We organize the experiments into two parts.
In Section{~\ref{sec:quadrant_expr}}, we implement different quadrants
in the same code base and assess their performance over a range of synthetic datasets.
In Section{~\ref{sec:end2end}}, we compare {\alg} with other baselines over extensive public and synthetic datasets.
For more experiments, including the efficiency of the horizontal-to-vertical
transformation and scalability of {\alg}, please refer to 
the appendix of our technical report~\cite{1907.01882}.

\subsection{Experimental Setup}
\textbf{Environment.}
We conduct the experiments on an 8-node laboratory cluster.
Each machine is equipped with 32GB RAM, 4 cores and 1Gbps Ethernet.
The maximum memory allowed for each run is limited to 30GB,
and we use 4 threads to achieve parallel computation on each node.

\textbf{Hyper-parameters.}
In specific experiments, we vary some hyper-parameters to assess
the change in performance. However, unless otherwise stated, 
we set $T=100$ (\# trees), $L=8$ (\# layers), and $q=20$ (\# candidate splits).

\begin{figure*}[!t]
	\centering
    \subfigure[{\small Impact of instance number. \newline D=100, C=2, L=8}]{
    \label{fig:dense_ins}
    \includegraphics[width=1.5in]{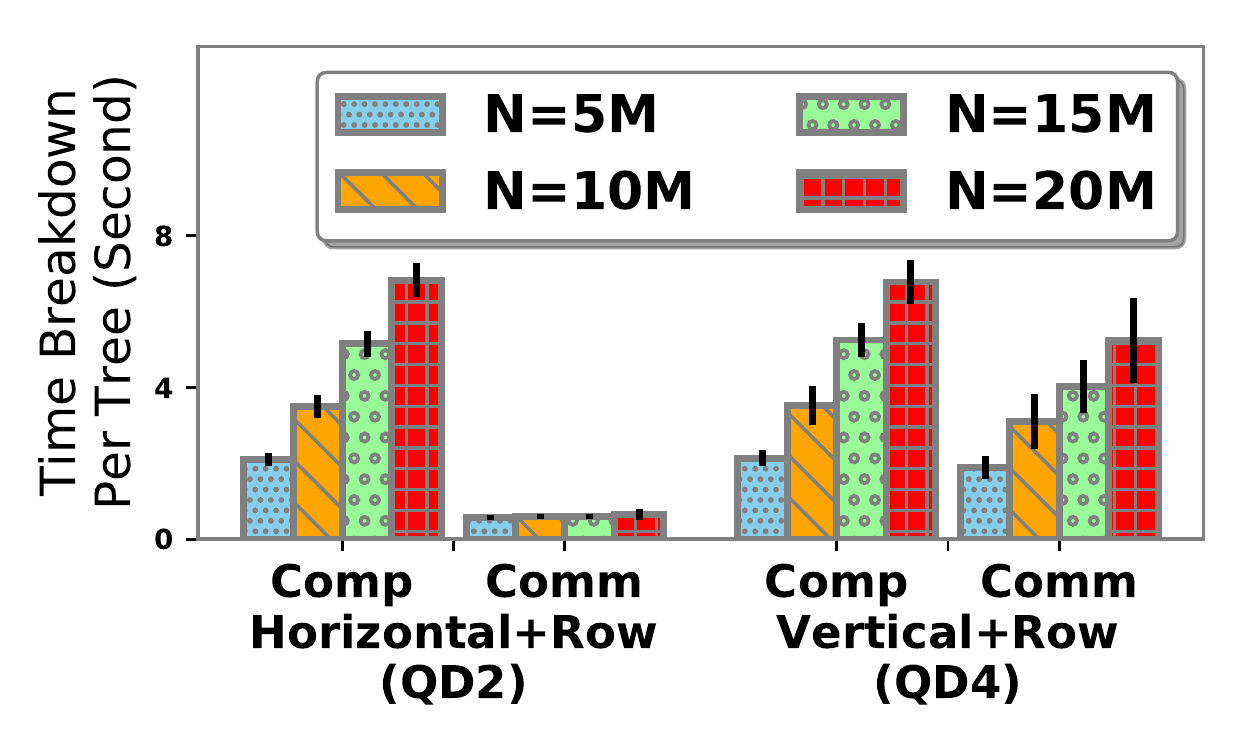}}
    \subfigure[{\small Impact of dimensionality. \newline N=50M, C=2, L=8}]{
    \label{fig:dimension}
    \includegraphics[width=1.5in]{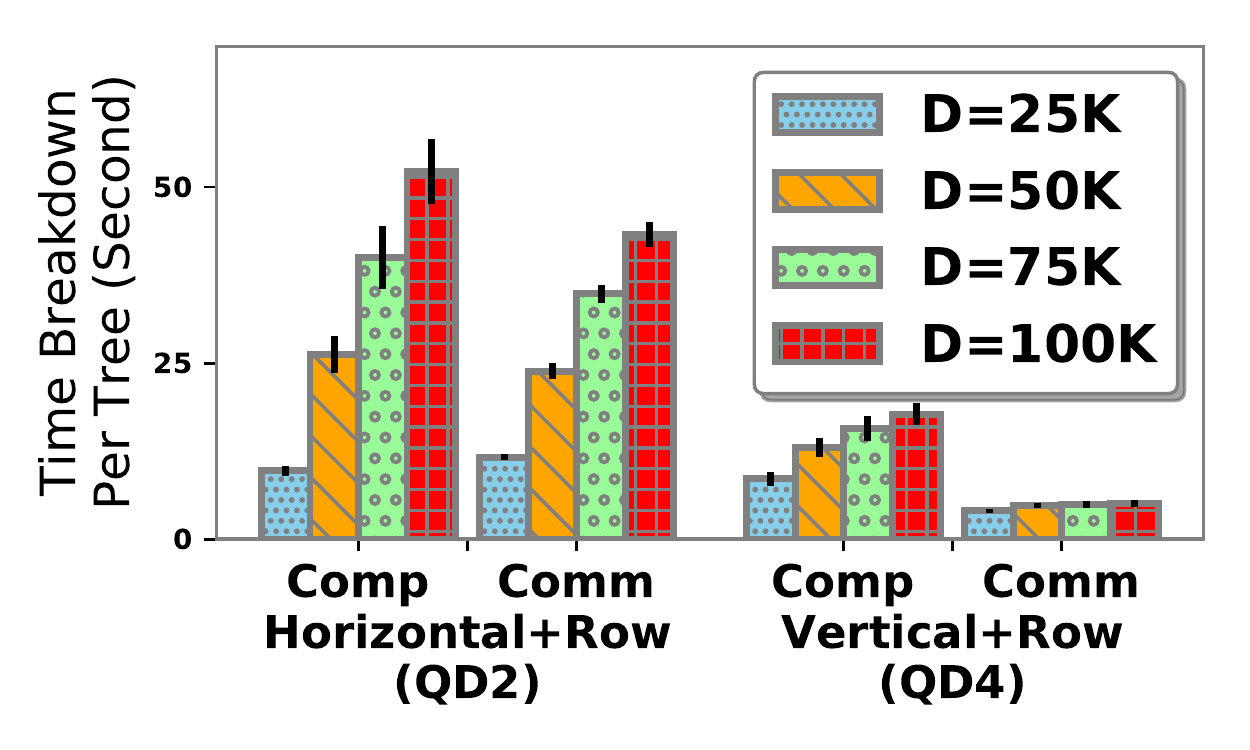}}
    \subfigure[{\small Impact of tree depth. \newline N=50M, D=100K, C=2}]{
    \label{fig:tree_depth}
    \includegraphics[width=1.5in]{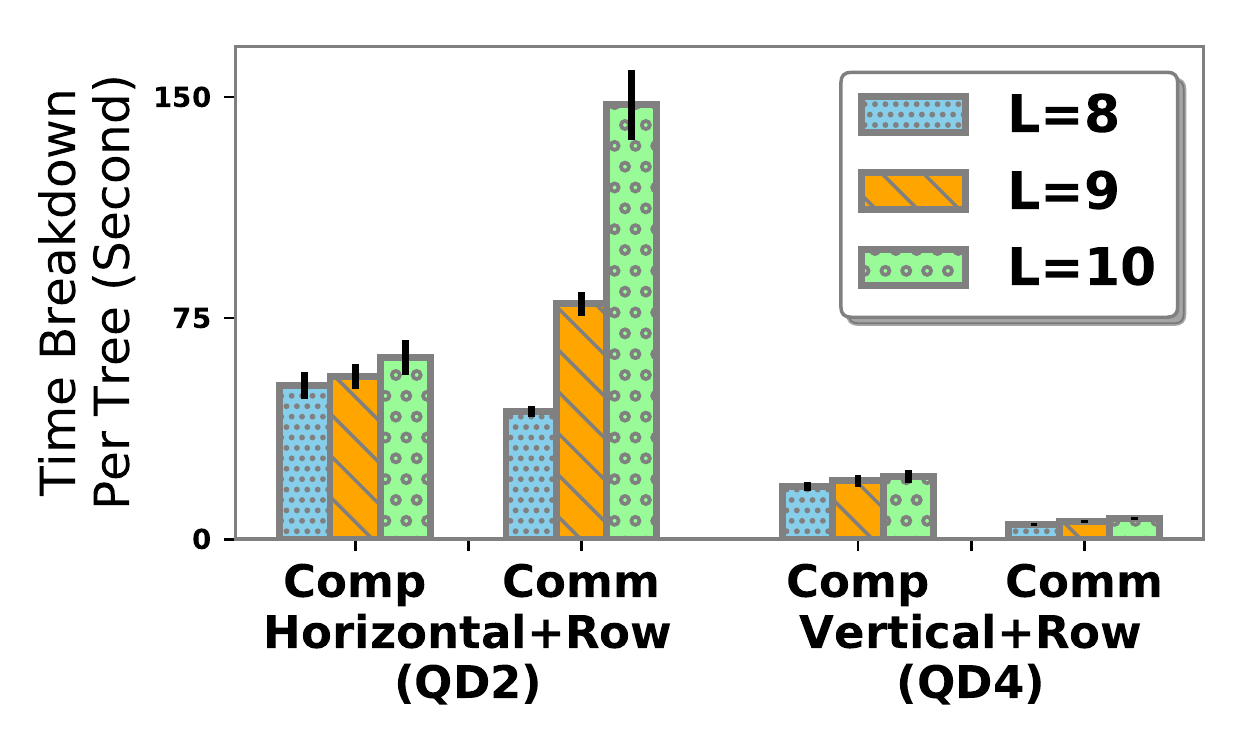}}
    \subfigure[{\small Impact of multi-classes. \newline N=50M, D=25K, L=8}]{
    \label{fig:multi-class}
    \includegraphics[width=1.5in]{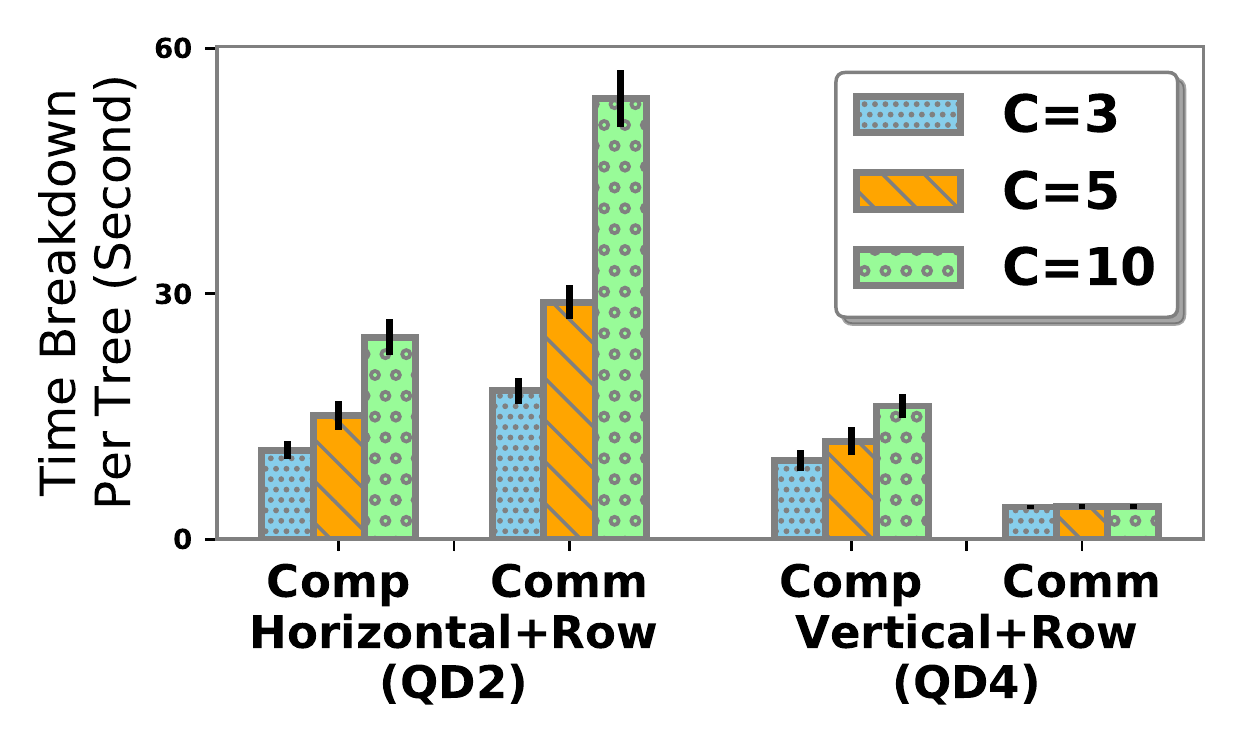}}
    \subfigure[{\small Memory consumption. \newline N=50M, C=2, L=8}]{
    \label{fig:memory_dim}
    \includegraphics[width=1.5in]{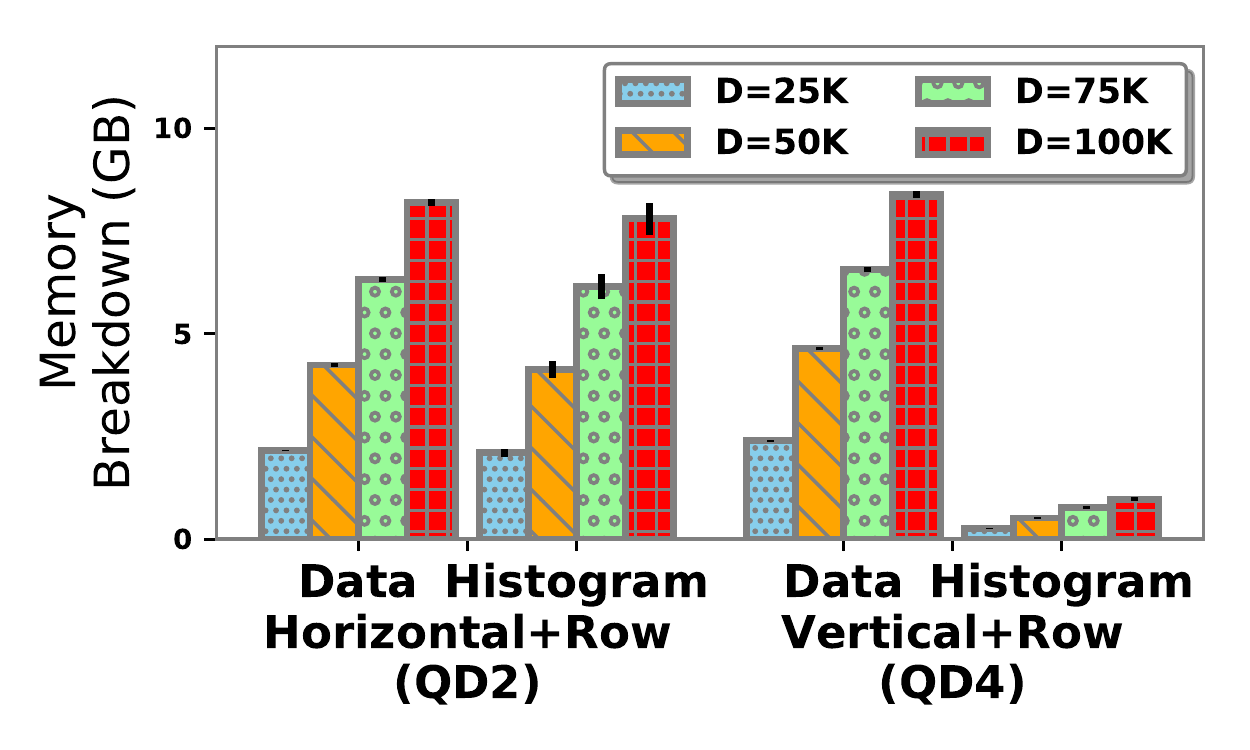}}
    \subfigure[{\small Memory consumption. \newline N=50M, D=25K, L=8}]{
    \label{fig:memory_multi}
    \includegraphics[width=1.5in]{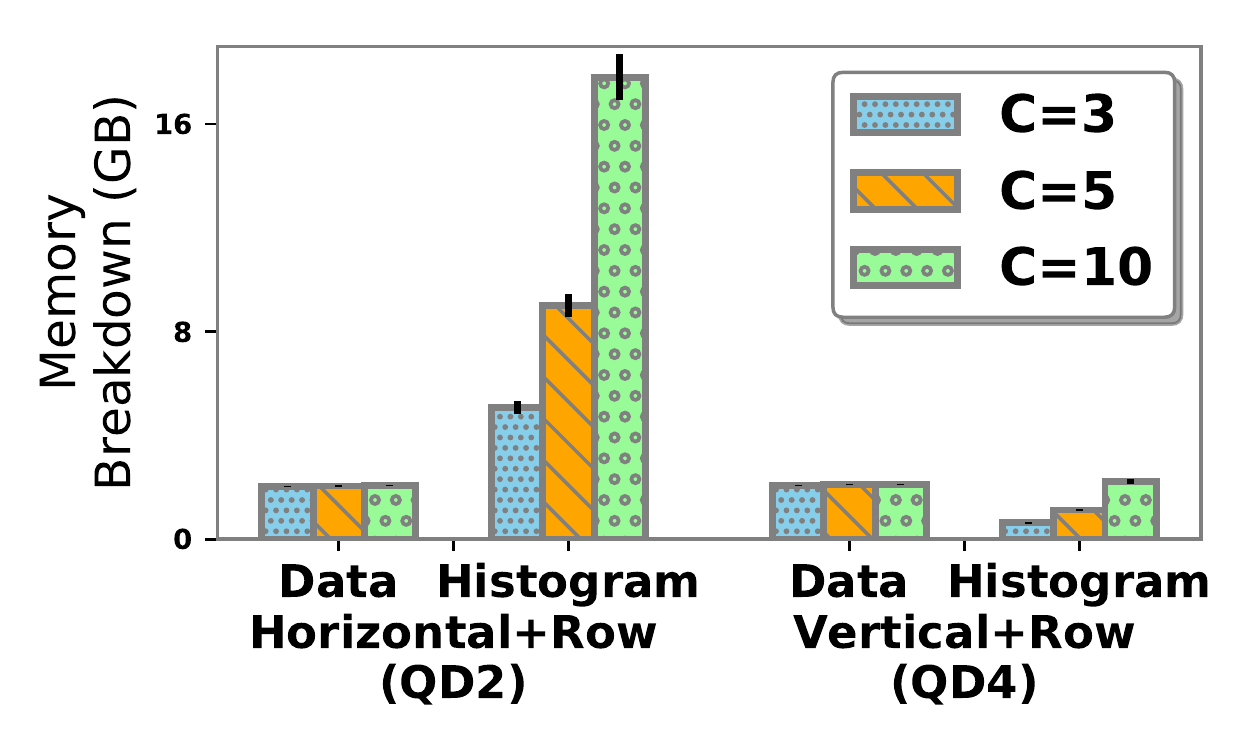}}
    \subfigure[{\small Impact of dimensionality. \newline N=10K, C=2, L=8}]{
    \label{fig:dim-storage}
    \includegraphics[width=1.5in]{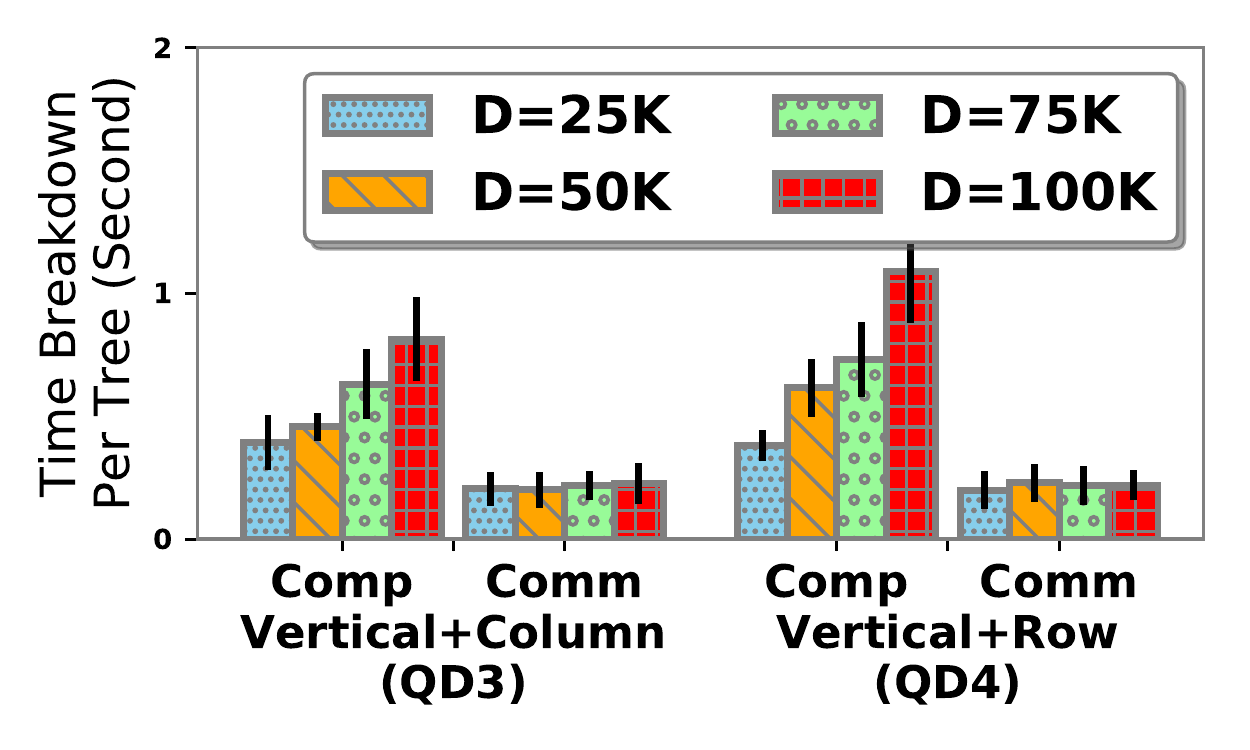}}
    \subfigure[{\small Impact of instance number. \newline D=100K, C=2, L=8}]{
    \label{fig:num_ins}
    \includegraphics[width=1.5in]{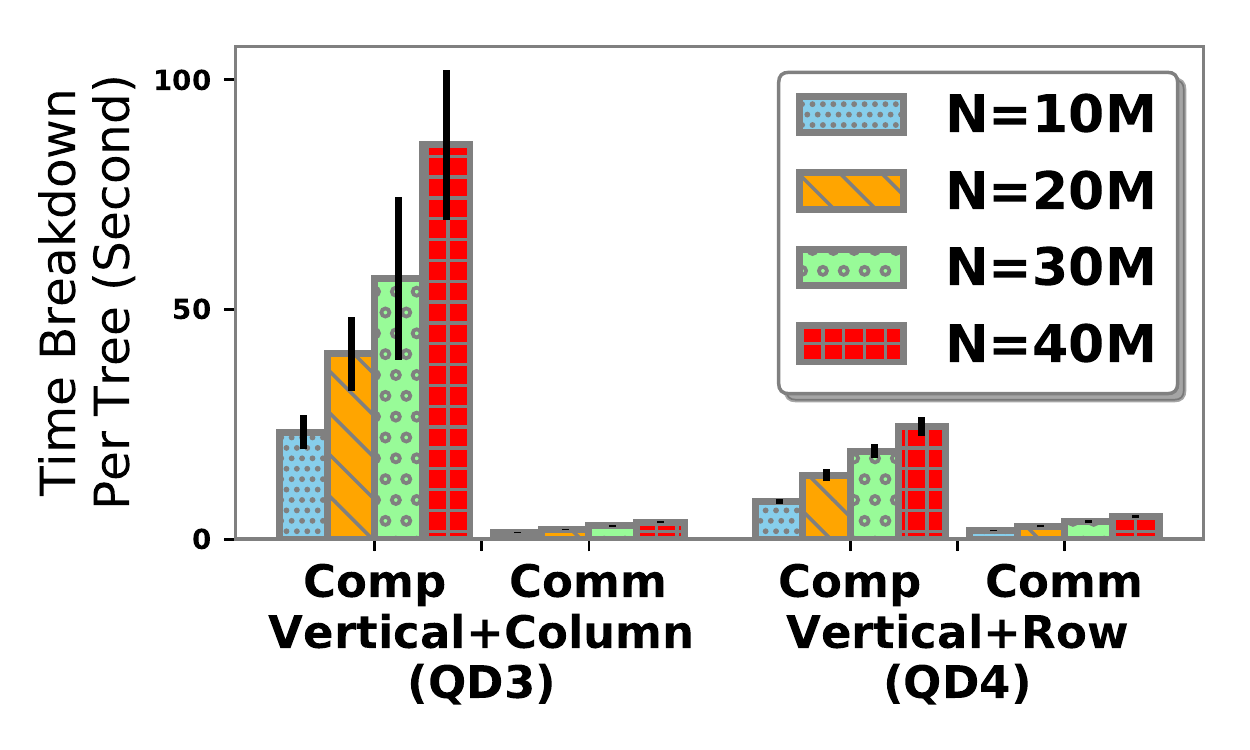}}
	\caption{Comparison of quadrants. Comp refers to computation, and Comm refers to communication.}
\end{figure*}

\subsection{Assessment of Quadrants}
\label{sec:quadrant_expr}
In order to validate the analysis in Section~\ref{sec:anatomy}, 
we evaluate the impact of partitioning scheme and storage pattern.
For partitioning scheme, we compare \alg with QD2, 
in terms of communication and memory efficiency.
For storage pattern, we compare \alg with QD3 in terms of computation efficiency.

To achieve fair and thorough comparison, 
we implement two optimized baselines in QD2 and QD3
on top of Spark and compare them with {\alg} over a range of synthetic datasets, 
and report the mean and standard deviation of one tree.
The synthetic datasets are generated from random linear regression models.
Specifically, given dimensionality $D$, informative ratio $p$, and number of classes $C$,
we first randomly initialize the weight matrix $W$ with size $D \times C$, and each row of $W$ contains $pD$ nonzero values.
Then for each instance, the feature $x$ is a randomly sampled $D$-dimensional vector with density $\phi$,
and its label $y$ is determined by $\argmax x^TW$.
In our experiment, we set $p=\phi=20\%$.

\subsubsection{Partitioning schemes}
\textbf{Impact of number of instances.}
We first assess the impact of number of instances $N$ using low-dimensional datasets, 
and present the time cost per tree in Figure{~\ref{fig:dense_ins}}.
The computation time of QD2 and QD4 is close to each other since partitioning scheme does not have influence on computation,
Nonetheless, the communication time varies. 
With $D=100$, which is a fairly low dimensionality, the communication cost of QD2 is negligible 
since the size of gradient histograms is small.
In contrast, QD4 takes nearly half of the training time on network transmission.
Besides, when $N$ grows larger, the communication cost of QD4 also becomes higher.
This is because vertical partitioning has to broadcast the placement of instances after node splitting,
which results in proportional network overhead w.r.t. $N$.
Therefore, given a low-dimensional datasets containing a large amount of instances,
horizontal partitioning is a properer choice.

\textbf{Impact of dimensionality.}
To assess the impact of feature dimensionality $D$, we train distributed GBDT over datasets with varying $D$,
as shown in Figure~\ref{fig:dimension}.
The communication time of horizontal partitioning increases linearly w.r.t. $D$, since the histogram size grows linearly,
while vertical partitioning gives almost the same communication time regardless of $D$.
The result validates that vertical partitioning is more communication-efficient for the high-dimensional datasets.
Theoretically speaking, the computation cost of QD2 and QD4 is similar, which matches the case when $D=25K$.
However, when we use more features, the computation time of QD2 increases sharply while that of QD4 grows mildly.
This is because when $D$ gets higher, the histogram becomes larger and cannot fit in cache.
Thus QD2 suffers from frequent cache miss, 
and therefore spends more time on histogram construction for larger $D$.
QD4, instead, holds a much smaller histogram on each worker owing to vertical partitioning and has a slow-growth in computation time.

\textbf{Impact of tree depth.}
We then assess the impact of the number of tree layers by changing $L$.
As shown in Figure~\ref{fig:tree_depth}, when $L$ increases from 8 to 9 and 10, 
the communication time of QD2 almost increases exponentially
because the number of tree nodes becomes exponential.
To the contrary, the communication time of QD4 increases linearly w.r.t $L$
since the transmission on each layer remains the same.
As for computation time, due to the histogram subtraction technique,
the time to build histograms for a deep layer is very little.
As a result, communication dominates when the decision tree goes deeper,
and vertical partitioning reveals its superiority more for deep trees.

\textbf{Impact of multi-classes.}
We next assess the impact of the number of classes $C$ in multi-classification tasks.
The experiments are conducted on several synthetic datasets with different number of classes.
Since QD2 encounters OOM (out-of-memory) error with $D=100K$ and $C=10$, we lower the dimensionality to 25K.
The results are presented in Figure~\ref{fig:multi-class}.
The computation time of QD2 and QD4 shows similar increase when $C$ increases from 3 to 5, and to 10.
Nevertheless, the communication time of QD2 is approximately proportional to $C$, while that of QD4 remains unchanged.
This validates our analysis that vertical partitioning is more suitable for multi-classification tasks than horizontal partitioning as it saves a lot of communication.

\textbf{Memory consumption.}
We record the memory consumption by monitoring the GC of JVM. 
As analyzed in Section~\ref{sec:anatomy}, the vertical partitioning is more memory-efficient since each worker does not need to store the histograms of all features.
Therefore, we breakdown the memory consumption into data and histogram.
As shown in Figure~\ref{fig:memory_dim} and Figure~\ref{fig:memory_multi}, 
QD2 and QD4 incur similar memory cost to store dataset.
QD4 allocates slightly more memory since it needs to store all instance labels.
Nonetheless, the memory for histogram is much different.
Compared to QD4, QD2 allocates approximately 6-8$\times$ space to persist the histograms, 
showing that the memory cost of vertical partitioning can be alleviated given more workers.
Moreover, in multi-classfication tasks, the memory consumption of histogram in QD2 dominates the overall memory cost, 
since the histogram size grows linearly against $C$ while the size of dataset remains unchanged.
QD4, to the contrary, is able to handle high-dimensional or multi-class datasets with limited memory resource.

\begin{figure*}[!t]
\begin{minipage}{.48\textwidth}
\scriptsize
\centering
\captionof{table}{\small Public and synthetic datasets. LD refers to low-dimensional dense datasets; HS refers to high-dimensional sparse datasets; MC refers to multi-classification datasets.}
\begin{tabular}{c c c c c c}
\toprule[1.5pt]
Dataset & Size & \# Ins & \# Feat & \# Labels & Type\\
\midrule[1pt]
SUSY & 2GB & 5M & 18 & 2 & LD \\
\midrule[1pt]
Higgs & 8GB & 11M & 28 & 2 & LD \\
\midrule[1pt]
Criteo & 10GB & 45M & 39 & 2 & LD \\
\midrule[1pt]
Epsilon & 15GB & 500K & 2K & 2 & LD \\
\midrule[1pt]
RCV1 & 1.2GB & 697K & 47K & 2 & HS \\
\midrule[1pt]
Synthesis & 60GB & 50M & 100K & 2 & HS \\
\midrule[1pt]
RCV1-multi & 0.8GB & 534K & 47K & 53 & MC \\
\midrule[1pt]
Synthesis-multi & 18GB & 50M & 25K & 10 & MC \\
\bottomrule[1.5pt]
\end{tabular}
\label{tb:pub_dataset}
\end{minipage}
\begin{minipage}{.04\textwidth}
\end{minipage}
\begin{minipage}{.48\textwidth}
\scriptsize
\centering
\captionof{table}{\small Average run time per tree scaled by {\alg}. We highlight the fastest ones in bold. \\ $ $}
\begin{tabular}{c c c c c}
\toprule[1.5pt]
Dataset & XGBoost & LightGBM & DimBoost & {\alg}\\
\midrule[1pt]
SUSY & 0.3 & \textbf{0.1} & 0.5 & 1.0  \\
\midrule[1pt]
Higgs & 0.5 & \textbf{0.2} & 0.8 & 1.0 \\
\midrule[1pt]
Criteo & 0.5 & \textbf{0.2} & 0.7 & 1.0 \\
\midrule[1pt]
Epsilon & 2.8 & \textbf{0.7} & 1.9 & 1.0 \\
\midrule[1pt]
RCV1 & 17.3 & 5.6 & 4.0 & \textbf{1.0} \\
\midrule[1pt]
Synthesis & 18.9 & 5.0 & 2.0 & \textbf{1.0} \\
\midrule[1pt]
RCV1-multi & 34.7 & 9.7 & - & \textbf{1.0} \\
\midrule[1pt]
Synthesis-multi & 7.1 & 3.3 & - & \textbf{1.0} \\
\bottomrule[1.5pt]
\end{tabular}
\label{tb:avg_time}
\end{minipage}
\end{figure*}


\begin{figure*}[!t]
    \centering
    \subfigure[{\small SUSY}]{
    \label{fig:susy}
    \includegraphics[width=1.6in]{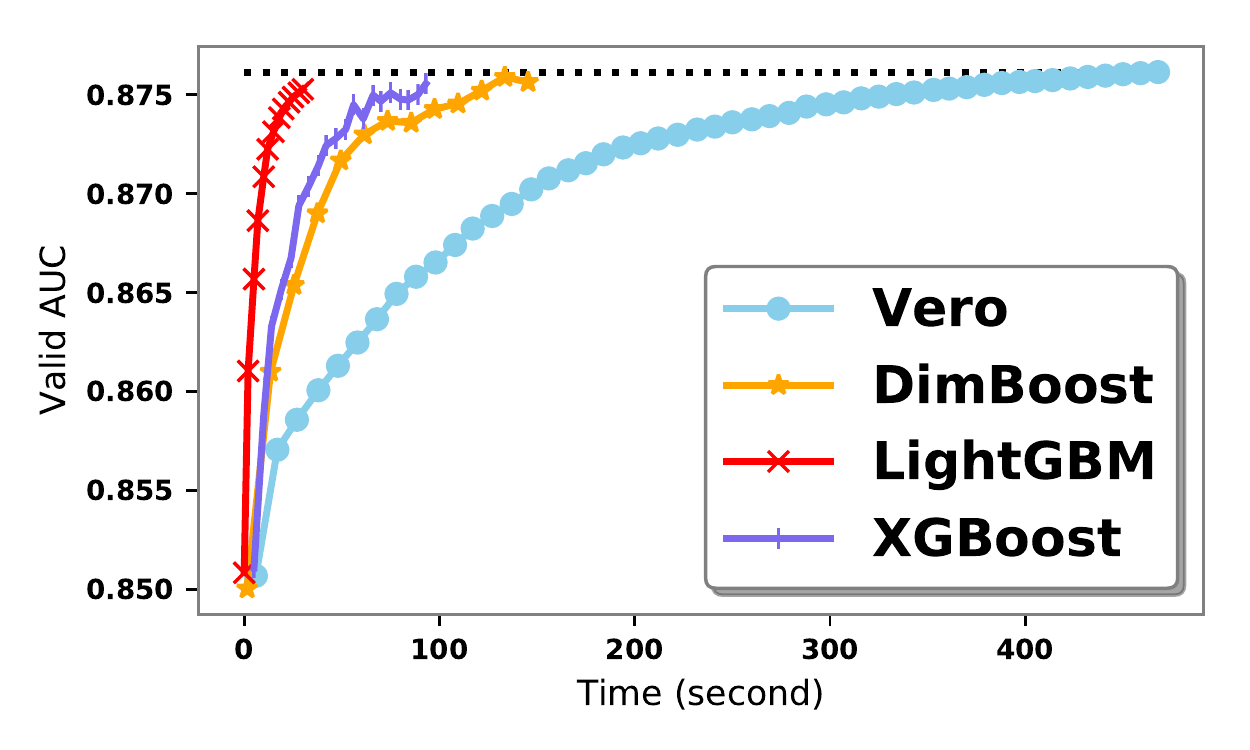}}
    \subfigure[{\small Higgs}]{
    \label{fig:higgs}
    \includegraphics[width=1.6in]{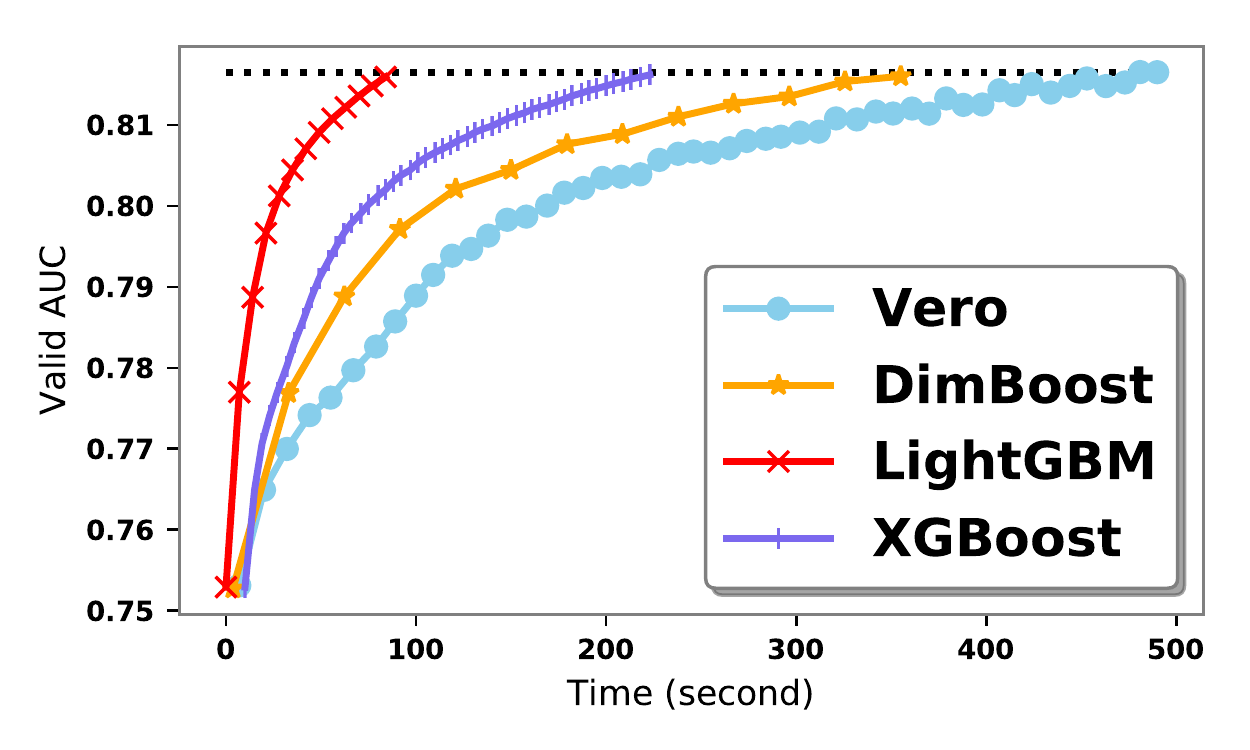}}
    \subfigure[{\small Criteo}]{
    \label{fig:criteo}
    \includegraphics[width=1.6in]{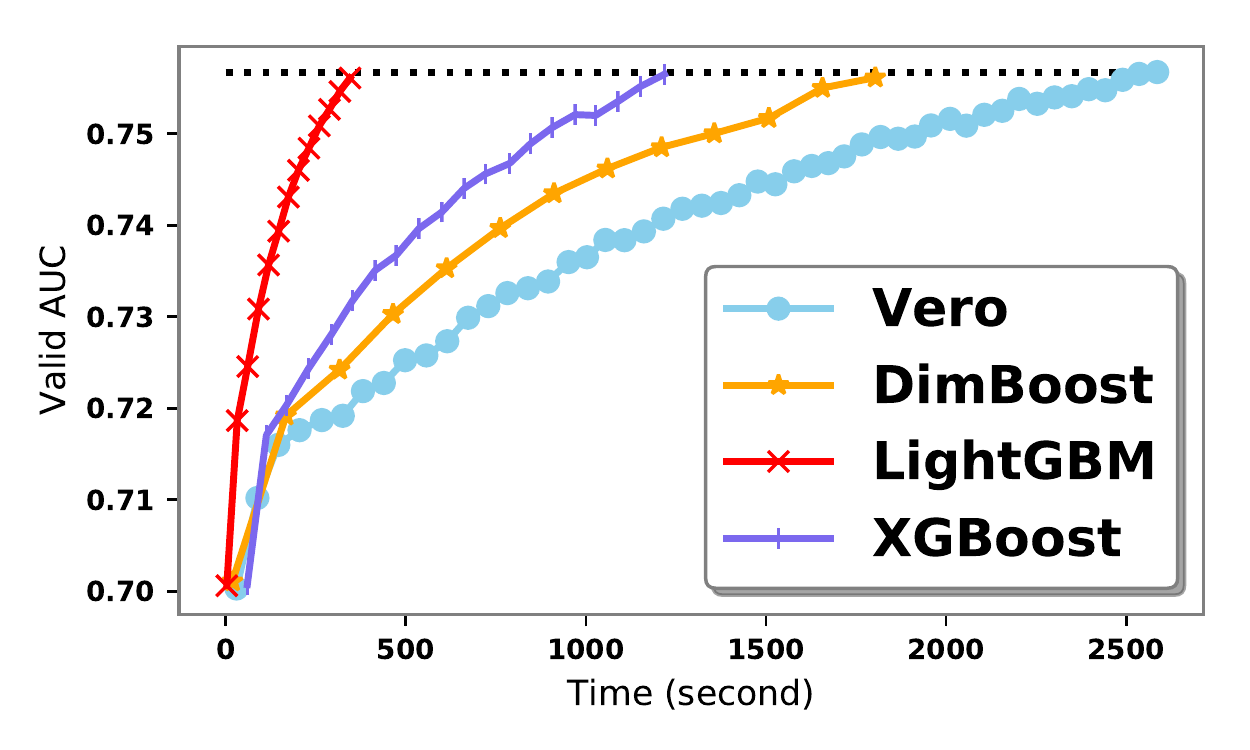}}
    \subfigure[{\small Epsilon}]{
    \label{fig:epsilon}
    \includegraphics[width=1.6in]{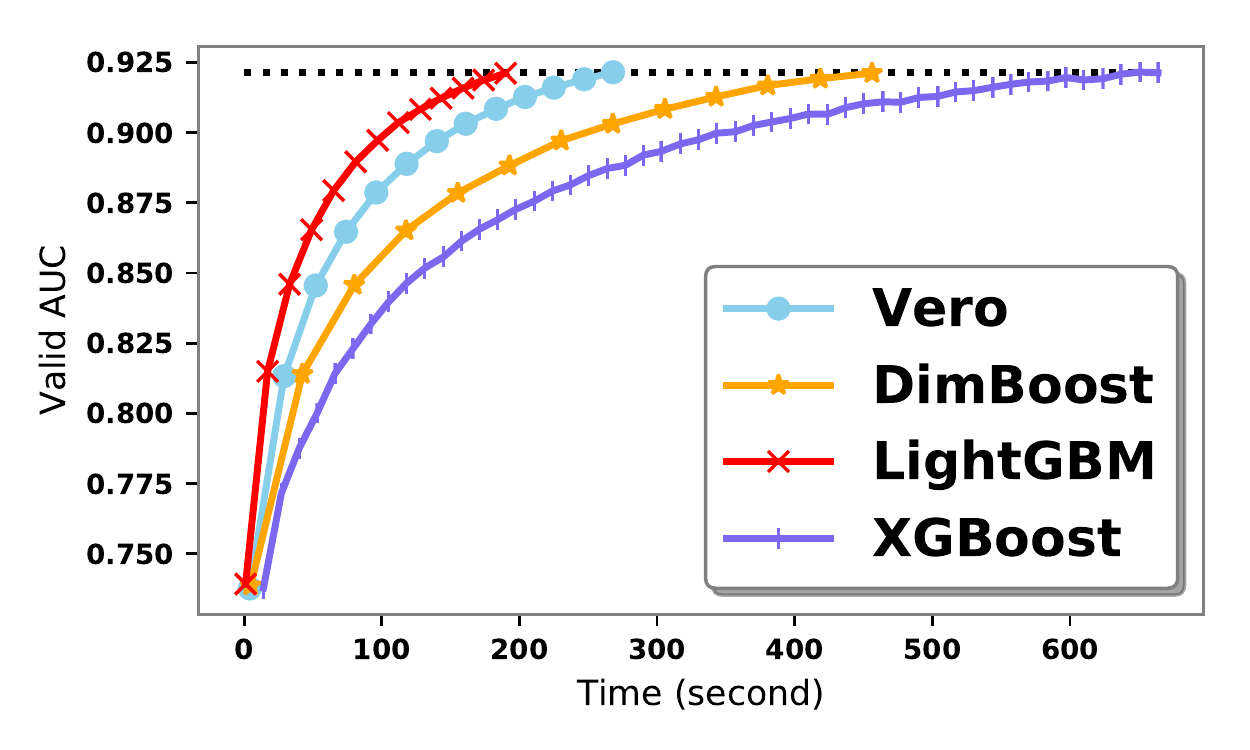}}
    \subfigure[{\small RCV1}]{
    \label{fig:rcv1}
    \includegraphics[width=1.6in]{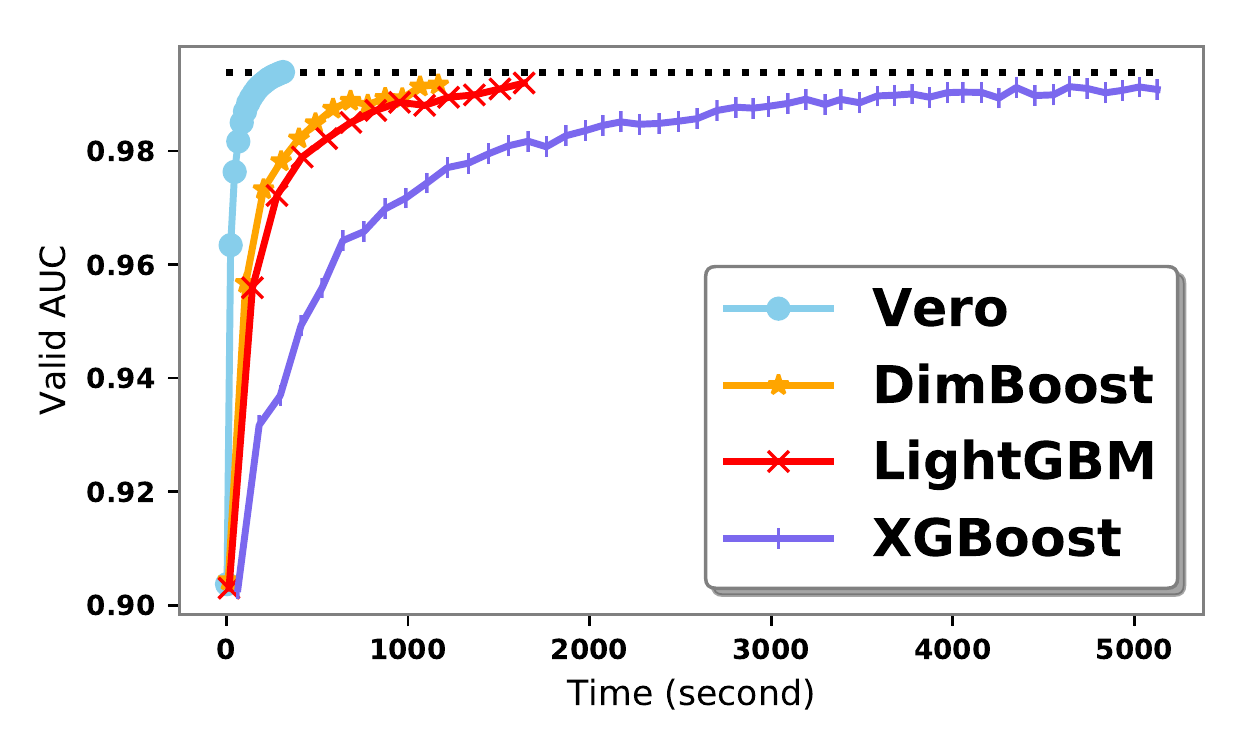}}
    \subfigure[{\small Synthesis}]{
    \label{fig:synthesis}
    \includegraphics[width=1.6in]{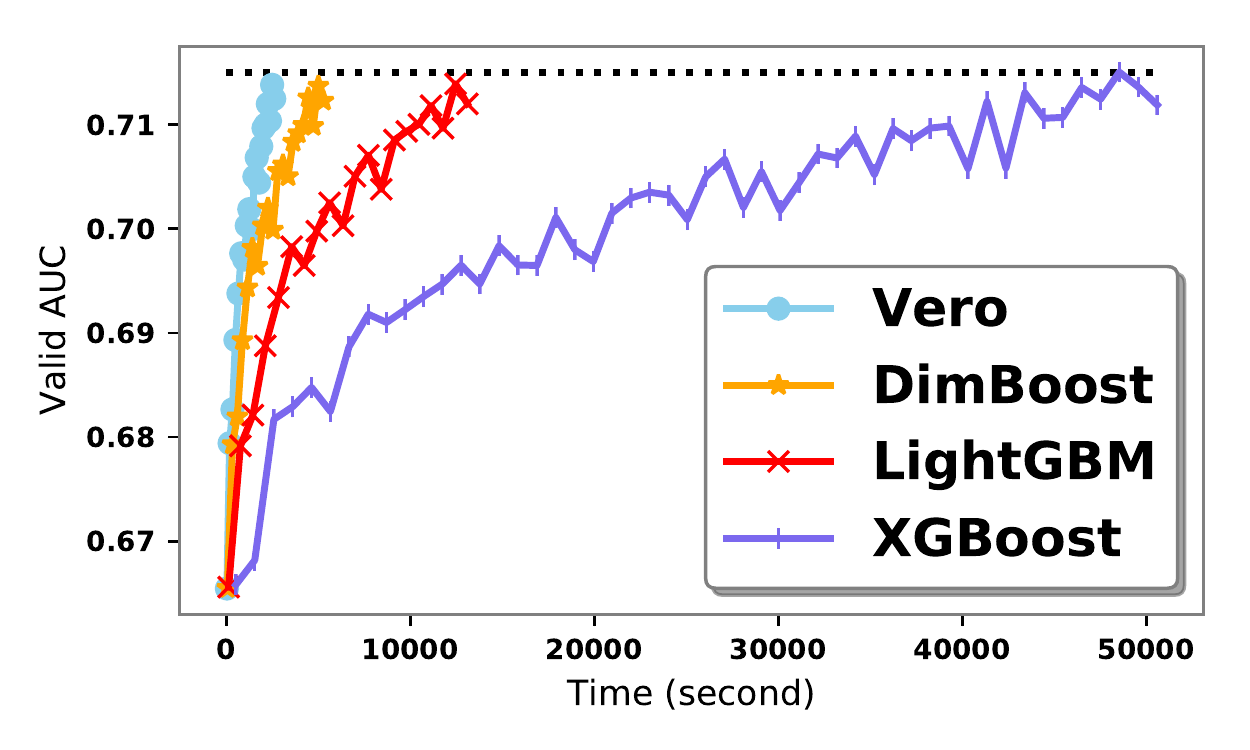}}
    \subfigure[{\small RCV1-multi}]{
    \label{fig:rcv1-multi}
    \includegraphics[width=1.6in]{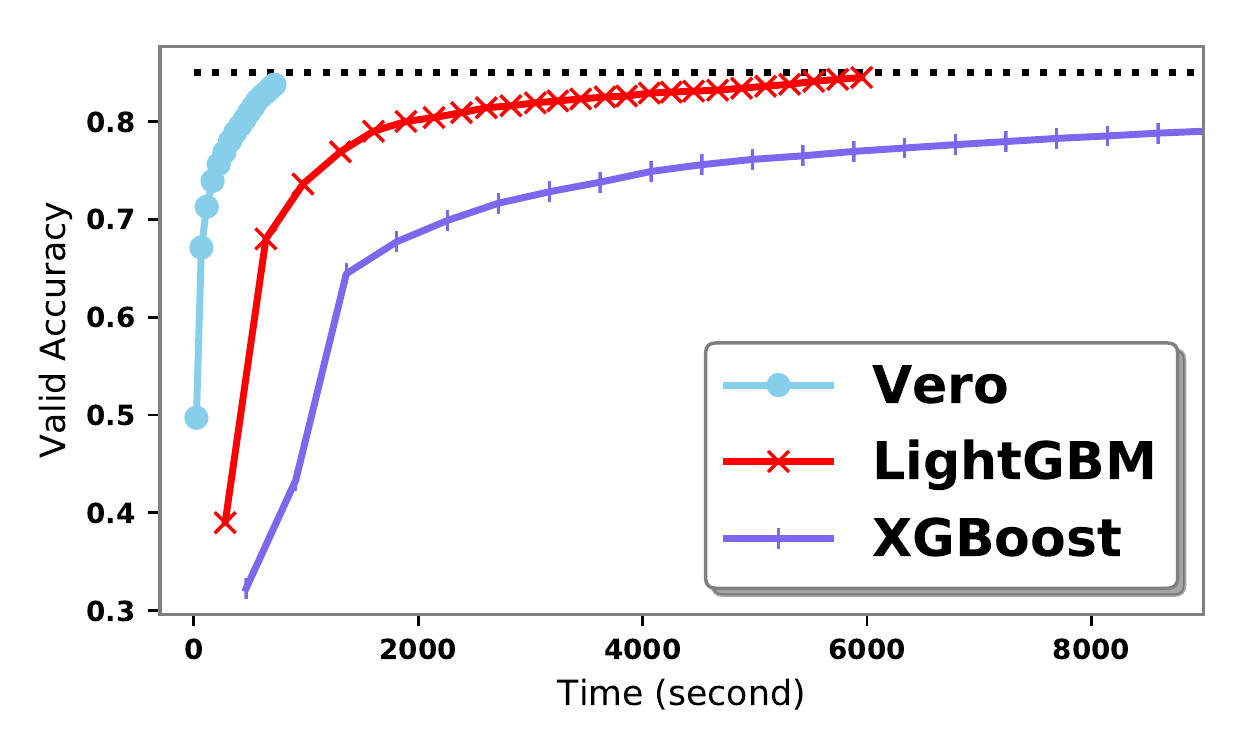}}
    \subfigure[{\small Synthesis-multi}]{
    \label{fig:synthesis-multi}
    \includegraphics[width=1.6in]{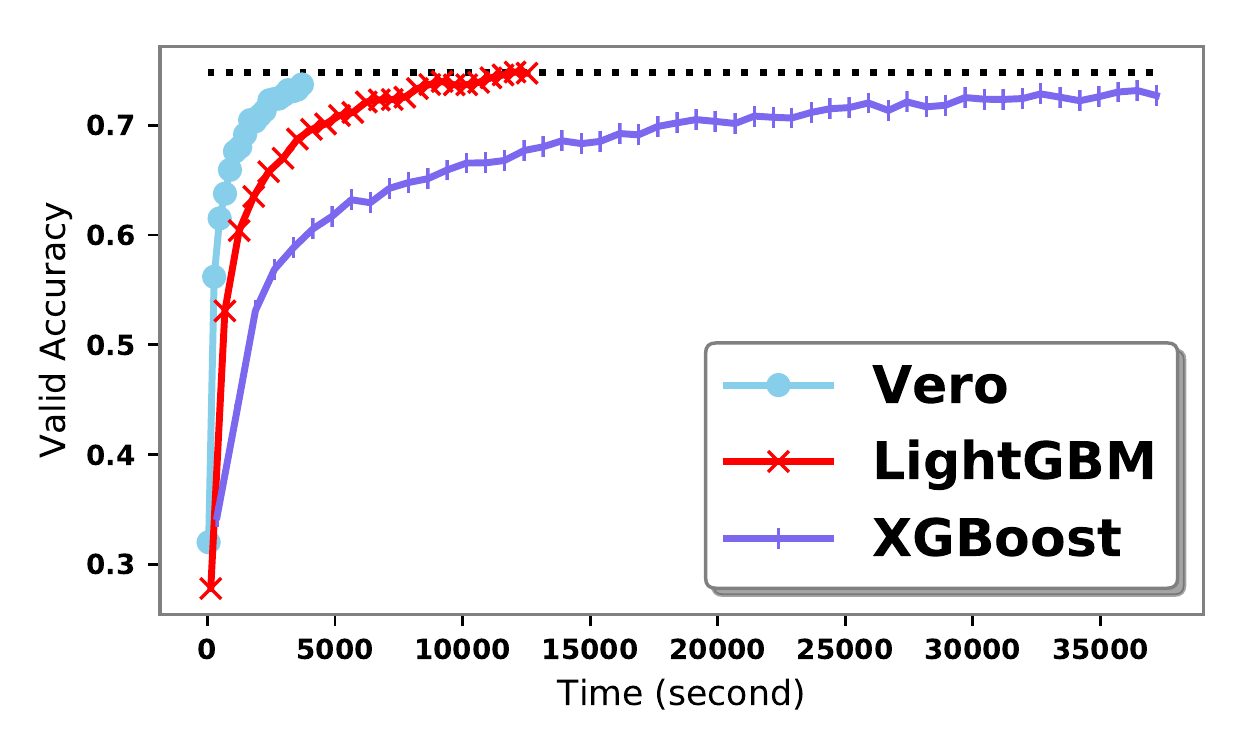}}
    \caption{\small{End-to-end evaluation. We report the convergence curves and draw a horizontal line to indicate the best model performance.}}
    \label{fig:pub_curve}
\end{figure*}

\subsubsection{Storage patterns}
\textbf{Index plan.}
Since the column-wise node-to-instance index causes unacceptable overhead during update, 
we implement QD3 with a combination of node-to-instance and instance-to-node indexes. Specifically, 
when a column contains few number of values, we build histogram for it by linear scanning, otherwise, we perform binary search on the column.
In the appendix of our technical report~\cite{1907.01882},
we compare our QD3 implementation with Yggdrasil to show that the combination of two indexes can achieve higher performance.

\textbf{Impact of dimensionality.}
We first study the performance on datasets with only a few instances but a high dimensionality.
Although such datasets are seldom seen in practice, conducting the comparison helps make our assessment complete.
The result is given in Figure{~\ref{fig:dim-storage}}.
Given a fixed $N$, the communication cost of QD3 and QD4 almost stays unchanged, due to the vertical partitioning they adopt.
However, QD4 spends more time on computation than QD3 given a larger $D$.
The reason is that QD3 stores the dataset column-by-column and constructs histograms one-by-one,
thus it is more cache-friendly when writing on the histograms.
While row-store constructs histograms for all features together, which will suffer from heavy cache miss when $D$ is large.
As a result, the experiment results match our analysis in Section{~\ref{sec:storage_pattern}} that 
column-store performs better than row-store
when the dataset is low-dimensional and meanwhile contains very few instances.

\textbf{Impact of number of instances.}
We then assess the impact of number of instances $N$. 
As shown in Figure~\ref{fig:num_ins}, 
QD3 and QD4 have similar network time growing linearly against $N$, 
since both of them vertically partition the datasets and need to transmit the instance placement.
The difference occurs in computation time.
In general, QD3 spends 3-4$\times$ on computation compared with QD4.
Moreover, the computation time of QD3 oscillates heavily (high standard deviation of time per tree).
This is because the binary searches on columns result in many CPU branch mispredictions.
In contrast, when training with row-store, we iterate the feature vectors row-by-row, 
which escapes from heavy branch prediction penalty.
In short, QD3 shares the same communication overhead of QD4,
but QD3 is not as computation-efficient as QD4, owing to the column-store it adopts.

\subsubsection{Summary}
The experiments above validate the analysis in Section~\ref{sec:anatomy}, that 
(i) horizontal partitioning works better when dimensionality is low, while
vertical partitioning is more memory- and communication-efficient 
under the high-dimensional, deep trees and multi-class cases; 
(ii) row-store is more efficient in computation than column-store 
except that the dataset is high-dimensional with few instances.
In addition, we observe another two advantages of QD4 in practice, which are cache- and branch-friendly.
As a result, the composition of vertical partitioning and row-store 
can achieve optimal performance under a wide range of workloads.

\subsection{End-to-end Evaluation}
\label{sec:end2end}
\textbf{Baselines.}
We choose three open source GBDT implementations as our baselines, which are XGBoost, LightGBM and DimBoost.
XGBoost and LightGBM are favorite toolkits in data-analytic competitions,
while DimBoost is optimized for large-scale GBDT workloads and is able to
achieve the state-of-the-art performance.

\textbf{Datasets.} 
We run {\alg} and the baselines on six public datasets and two synthetic datasets,
as listed in Table{~\ref{tb:pub_dataset}}.
We categorize the datasets into low-dimensional dense (LD), high-dimensional sparse (HS), and multi-classification (MC) datasets,
and discuss the overall performance of the systems on different kinds of datasets.
All systems are tuned to achieve comparable accuracy.
We present the convergence curve in Figure{~\ref{fig:pub_curve}} 
and report the running time in Table{~\ref{tb:avg_time}}.

\paragraph*{Low-dimensional Dense Datasets}
We first conduct end-to-end evaluation on four datasets  with low dimensionality and fully dense data.
We use five workers to run on these four datasets.
Corresponding to the analysis in Section{~\ref{sec:anatomy}},
low dimensionality results in small histogram size and hence the 
communication time of horizontal partitioning does not dominant.
Therefore, LightGBM, which belongs to QD2, achieves the fastest speed in overall,
since it is more computation-efficient than XGBoost (QD1) and 
communicates little compared to {\alg} (QD4).
{\alg} suffers on extreme low-dimensional datasets, i.e., 
\textit{SUSY}, \textit{Higgs}, and \textit{Criteo},
however, it catches up quickly and is comparable to LightGBM when the dimensionality gets higher, for instance the \textit{Epsilon} dataset, 
which also matches our analysis.
DimBoost (QD2) runs slower than XGBoost on three datasets, violating our analysis.
The unsatisfactory performance of DimBoost is caused by two factors: 
1) DimBoost is designed aiming at the high-dimensional case 
and always stores datasets as sparse matrices, 
which inevitably results in extra cost in data access and indexing;
2) DimBoost is implemented in Java, thus it is hard to achieve as good computation efficiency as 
the C++-based XGBoost and LightGBM.

\paragraph*{High-dimensional Sparse Datasets}
We then assess the systems on high-dimensional sparse datasets,
\textit{RCV1} and \textit{Synthesis},
with five and eight workers, respectively.
In short, {\alg} runs the fastest, followed by DimBoost and LightGBM, while XGBoost is the slowest.
XGBoost is about 18$\times$ slower than {\alg}, due to the inefficiency in both computation and communication.
The speedup of {\alg} w.r.t. DimBoost and LightGBM are 2-5.6$\times$.
The relative performance of {\alg} on \textit{Synthesis} is slower than \textit{RCV1},
since there is a large number of instances compared with the 330 thousand feature.
However, it can still achieve the fastest speed, owing to the superiority of QD4 under high-dimensional cases.

\paragraph*{Multi-classification Datasets}
Finally we consider the performance on multi-classification datasets using eight workers.
Since DimBoost does not support multi-classification,
we do not discuss it in this experiment.
XGBoost and LightGBM are 8.6$\times$ and 7.4$\times$ slower on the multi-class dataset \textit{RCV1-multi} than the binary-class dataset \textit{RCV1},
due to the 53$\times$ increment in network transmission.
{\alg}, however, takes only 4$\times$ more time on \textit{RCV1-mutli},
since the network transmission of vertical partitioning does not increase w.r.t. the number of classes.
Overall, {\alg} is 9.7$\times$ and 34.7$\times$ faster than LightGBM and XGBoost.
The speedup of {\alg} on \textit{Synthesis-multi} is smaller than \textit{Synthesis} 
due to the lower dimensionality, however, {\alg} still outperforms XGBoost and LightGBM 
by 7.1$\times$ and 3.3$\times$, respectively.
The experiment results match our analysis that QD4 is more suitable for multi-classification tasks.

\paragraph*{Summary}
The end-to-end evaluation reveals that we should choose the proper system for a given workload. 
To summarize, LightGBM achieves the highest performance on low-dimensional datasets, 
while {\alg} is the best choice for high-dimensional or multi-classification datasets.

\section{Evaluation in the Real World}
\label{sec:industrial}

As aforementioned, \alg has been integrated into the production pipeline of Tencent.
In this section, we present some use cases to validate the ability of \alg to
handle large scale real-world workloads.

\textbf{Environment.}
The experiments are carried out on a productive cluster in Tencent.
Each machine is equipped with 64GB RAM, 24 cores and 10Gbps Ethernet.
Since the cluster is shared by other applications, the maximum resource for 
each Yarn container is restricted.
Thus we use 20GB memory and 10 cores for each container.

\textbf{Datasets.}
We use three datasets in Tencent.
All three datasets are used to train models to complete the user persona.
\textit{Gender} contains 122 million instances.
\textit{Age} classifies 48 million users into 9 age ranges.
Both of them have 330 thousand features.
\textit{Taste}, with 10 million instances and 15 thousand features, describes the user taste with 100 tags.


\textbf{Hyper-parameters.}
We use 50 workers for \textit{Gender}, 20 workers 
for \textit{Age} and \textit{Taste}.
We set $T=20$ (\# trees) and restrict the maximum running time to convergence as 1 hour.
The other hyper-parameters are the same as in Section~\ref{sec:eval}.

\textbf{Baselines.}
Prior to \alg, XGBoost and DimBoost are two candidates for GBDT in Tencent.
As discussed in~\cite{jiang2018dimboost}, LightGBM
is impractical for productive environments owing to
the strict environment requirement and the lack of integration with the Hadoop ecosystem.
Therefore, we choose XGBoost and DimBoost as our baselines in this section.

\begin{figure*}[!t]
\begin{minipage}{.7\textwidth}
\centering
\includegraphics[width=1.6in,trim={0 0 4.95in 0},clip]{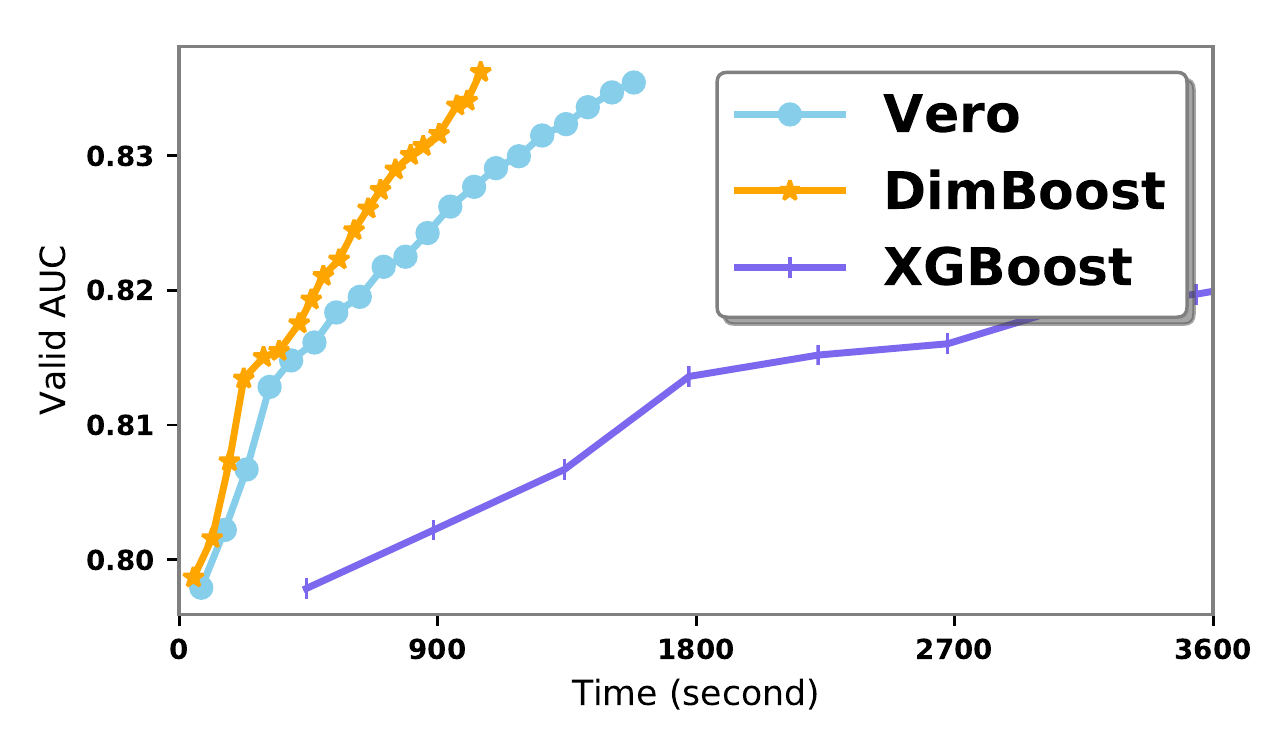}
\includegraphics[width=1.6in,trim={0 0 4.95in 0},clip]{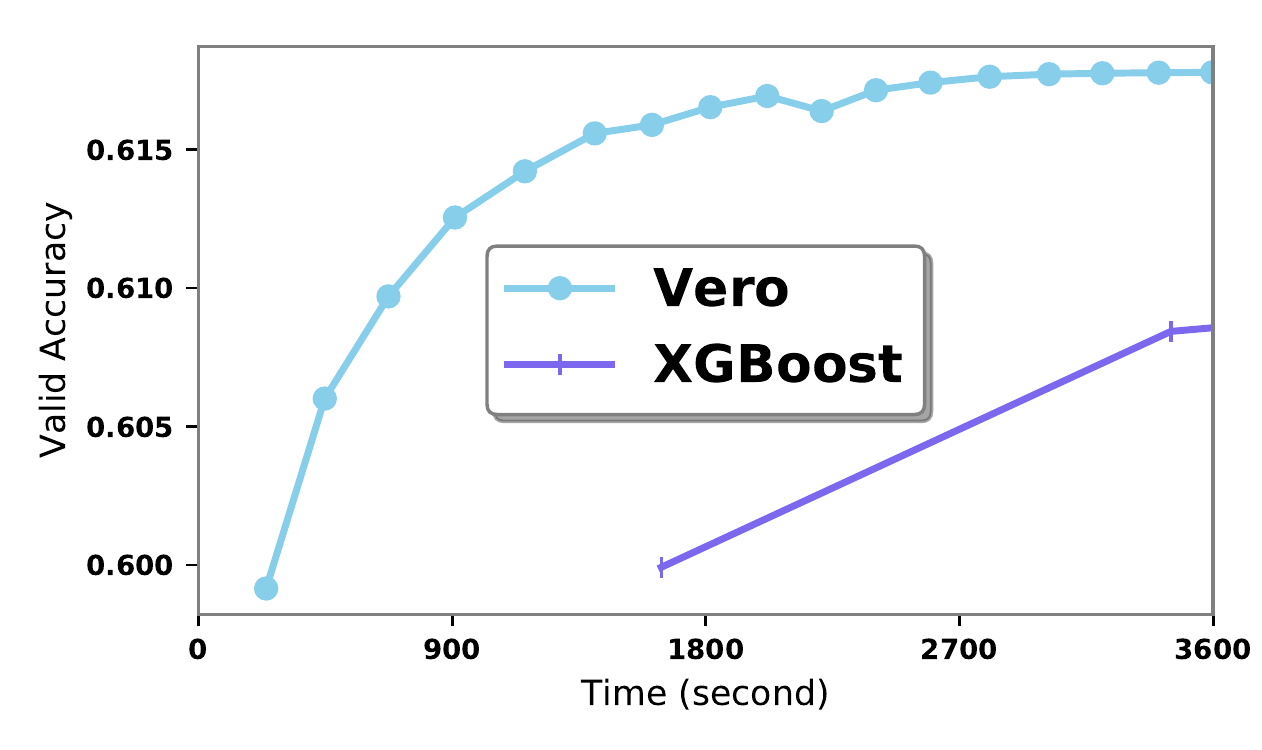}
\includegraphics[width=1.6in,trim={0 0 4.95in 0},clip]{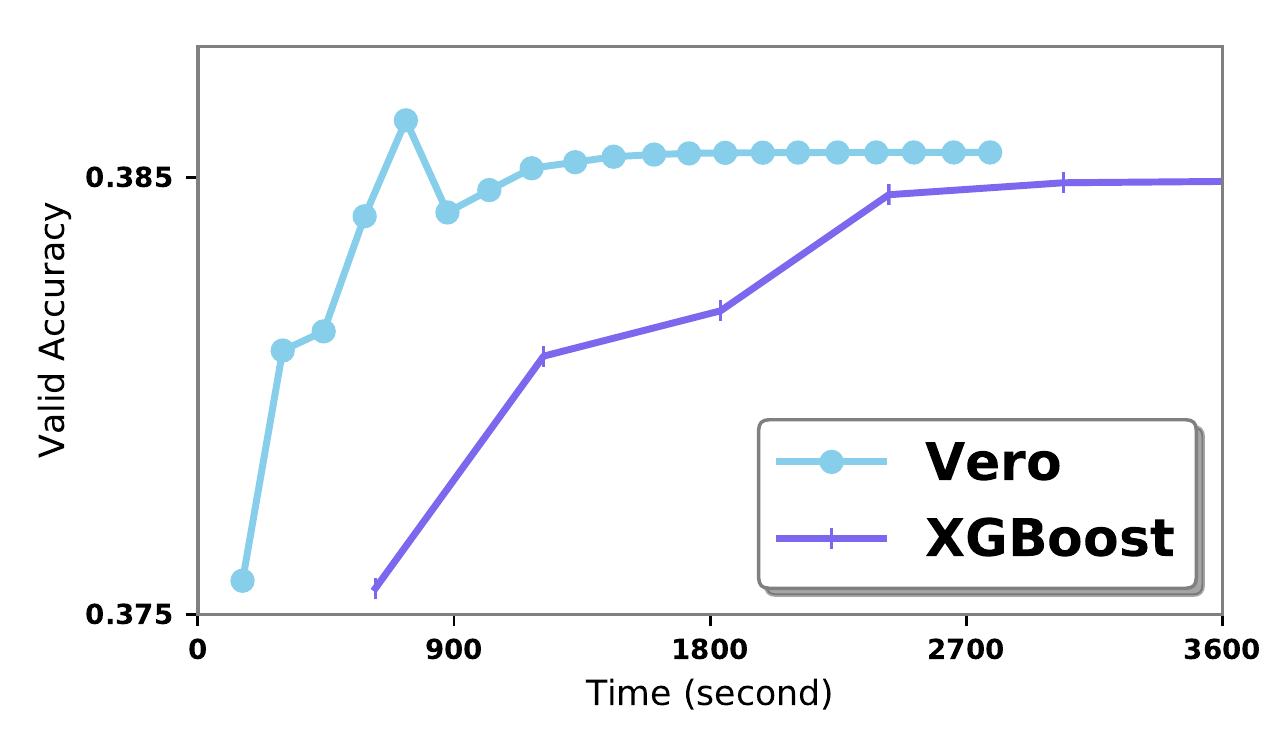}
\captionof{figure}{\small{End-to-end evaluation over industrial datasets. (left to right: Gender, Age, Taste)}}
\label{fig:industrial_curves}
\end{minipage} \quad
\begin{minipage}{.25\textwidth}
\centering
\scriptsize
\captionof{table}{\small Run time per tree in seconds (fastest ones in bold).}
\label{tb:industrial_time}
\begin{tabular}{c c c c}
\toprule[1.5pt]
Dataset & Gender & Age & Taste \\
\midrule[1pt]
XGBoost & 438 & 1738 & 627  \\
\midrule[1pt]
DimBoost & \textbf{52} & - & - \\
\midrule[1pt]
{\alg} & 79 & \textbf{207} & \textbf{139} \\
\bottomrule[1.5pt]
\end{tabular}
\end{minipage}
\end{figure*}


\paragraph*{Gender dataset}
We run the \textit{Gender} dataset on all the three systems and present the results in Figure~\ref{fig:industrial_curves} and Table~\ref{tb:industrial_time}.
Unfortunately, \alg spends 1.5$\times$ to finish one tree compared with DimBoost. 
This is caused by two factors. 
First, the productive cluster has a 10$\times$ higher network bandwidth compared to the laboratory cluster in Section~\ref{sec:eval}, so the communication overhead is alleviated for DimBoost.
Second, \textit{Gender} contains an extreme large amount of instances, in which case horizontal partitioning can better distribute the workloads to workers.
However, the time cost of \alg is comparable to that of DimBoost and can outperform XGBoost by 5.5$\times$, 
verifying that \alg can well support datasets with large number of instances and low dimensionality.

\paragraph*{Age dataset}
We next assess the performance of \alg and XGBoost on the large-scale multi-class dataset.
Figure~\ref{fig:industrial_curves} and Table~\ref{tb:industrial_time} give the results.
It takes 207 seconds for \alg to complete one tree,
and it can get close to convergence within an hour.
Nevertheless, XGBoost costs 1738 seconds for one tree, which is 8.3$\times$ slower.
In many real applications, the allowed time is usually restricted.
For instance, daily recurring jobs need to commit within a reasonable period of time so that the jobs in downstream will not be affected.
Obviously, XGBoost fails to converge within acceptable time on this dataset,
whereas \alg can achieve better performance since it is more efficient in both communication and computation.

\paragraph*{Taste dataset}
Finally we conduct an experiment on a relatively small-scale multi-class dataset.
As shown in Figure~\ref{fig:industrial_curves} and Table~\ref{tb:industrial_time}, \alg is 4.5$\times$ faster than XGBoost.
Although the feature dimensionality of \textit{Taste} is low, \alg can still outperform XGBoost, showing that \alg is more suitable for the multi-classification tasks. 

\paragraph*{Summary}
With the experimental results on three industrial datasets, we show that 
by careful investigation on the management of distributed datasets, we can achieve a better solution to solve a wide range of workloads.
Currently \alg is designed for vertical partitioning and row-store, and is not able to achieve highest performance on all cases.
How to determine an optimal dataset management strategy given the size of dataset (e.g., number of instances, feature dimensionality and number of classes) 
along with the application environment (e.g., network bandwidth, number of machines, number of cores)
is remained unsolved.
We believe this problem can bring insight to both the machine learning and database community and leave it as our future work.

\section{Related Work}
\label{sec:related}

A lot of works have implemented the algorithm, either in research interests or industrial needs. 
R-GBM and scikit-learn~\cite{ridgeway2007generalized,pedregosa2011scikit} 
are stand-alone packages so that they cannot handle large-scale datasets.
MLlib~\cite{meng2016mllib,zhang2019mllib} is a machine learning package of Spark and implements GBDT.
XGBoost~\cite{chen2016xgboost} achieves great success in various data analytics competitions,
and is also widely-used in companies due to the 
distributed learning supported by DMLC.
LightGBM~\cite{ke2017lightgbm} is developed in favor of data analytics.
Although it supports parallel learning with MPI, LightGBM requires complex setup
and is not a good fit for large scale workloads in commodity environment.
Note that there is a feature-parallel version of LightGBM, which lets each worker process a feature subset like vertical partitioning does. 
However, it requires all workers to load the whole dataset into memory, i.e. dataset is never partitioned, which is impractical for large-scale workloads. 
In Appendix of our technical report~\cite{1907.01882}
we conduct experiments on small datasets with the feature-parallel LightGBM and \alg. 
There is a surge of interests to introduce parameter-server architecture into 
industrial applications~\cite{jiang2017angel,zhou2017kunpeng,zhang2019ps2}. 
Notably, TencentBoost and PSMART~\cite{TencentBoost,zhou2017psmart} implement GBDT with parameter-server.
DimBoost~\cite{jiang2018dimboost} further applies a series of optimization techniques
and achieves the state-of-the-art performance.
However, it only supports binary-classification.

There exist many works discussing the impact on databases brought by data layout.
Column-oriented databases~\cite{stonebraker2005c,abadi2006integrating} vertically 
partition the data and store them in columns and outperform row-oriented
databases on database analytics workloads. ~\cite{abadi2008column} discusses the
performance difference in terms of row-store and column-store.
There are also works that take advantages of both vertical partitioning and row representation
~\cite{agrawal2004integrating,cui2010exploring}.
Despite the extensive studies in database community, how does the way we manage
the training datasets influence the performance of machine learning algorithms 
is few discussed.
Yggdrasil~\cite{abuzaid2016yggdrasil} introduces vertical partitioning into the training
of decision tree and showcases the reduction in network communication. 
Our work extends the analysis to both communication and memory overhead.
In addition, Yggdrasil focuses on the case of deep decision tree.
We further show that vertical partitioning combined with row-store benefits the 
high-dimensional and multi-classification cases.
DimmWitted~\cite{zhang2014dimmwitted} analyzes the trade-off in access methods
when training linear models under the NUMA architecture.
However, instances are stored in row format without vertical partitioning in DimmWitted.
In this work, we together discuss the data access and data index methods
for both row-store and column-store data when training GBDT.

The analysis in this work is applicable to many other tree-based algorithms beyond GBDT,
such as AdaBoost, random forest, and gcForest~\cite{freund1997decision,breiman2001random,zhou2017deep}.
However, there are also algorithms that our analysis fails to support.
For instance, neural decision forest~\cite{rota2014neural,kontschieder2015deep}
utilizes neural networks
(randomized multi-layer perceptron or fully-connected layers concatenated with a deep convolutional network)
as splitting criteria.
There is a big difference between this algorithm and vanilla decision trees.
To discuss the impact on performance brought by data management methods, 
we need thorough investigation on deep neural network training,
such as the anatomy of data parallelism and model parallelism.
Moreover, the qualitative study on how hardware environment
influences the performance is remained undone.
We leave these as future works and do not discuss them in this work.

\section{Conclusion}
\label{sec:conclusion}

In this paper, we systematically study the data management methods in distributed GBDT.
Specifically, we propose the four-quadrant categorization along partitioning scheme and storage pattern,
analyze their pros and cons, 
and summarized their advantageous scenarios in Table~\ref{tb:adv_summary}.
Based on the findings, we further propose {\alg}, a distributed GBDT implementation 
that partitions the dataset vertically and stores data in row manner.
Empirical results on extensive datasets validate our analysis and 
provide suggestive guidelines on choosing a proper platform for a given workload.

\textbf{Acknowledgements}. 
Jiawei Jiang is the corresponding author. 
This work is supported by the National Key Research and Development Program of China (No. 2018YFB1004403), NSFC(No. 61832001, 61702015, 61702016, 61572039), and PKU-Tencent joint research Lab.


\balance


\bibliographystyle{abbrv}
\bibliography{reference}

\ifvldb
\else
\newpage
\appendix
\section{Efficiency of Transformation}
\label{sec:transform}
We first study the efficiency of our horizontal-to-vertical transformation algorithm.
We show the time cost of data loading, candidate split finding, label broadcasting and horizontal-to-vertical repartition in Table~\ref{tb:transform_expr}.

\textbf{Effects of proposed techniques.}
To access the effects of individual optimizations,
we also implement the na\"ive method that transmits original 12-byte key-value pairs and 
a compression method that compresses key-value pairs without the blockify technique.
The results show that our algorithm can complete transformation with minimal time cost.
Taking \textit{Synthesis} as an example, the compression technique brings a 16\% reduction in time, 
and the blockify technique brings another 42\%.

\textbf{Analysis of transformation overhead.}
Note that both horizontal and vertical partitioning need to calculate data sketches (calculate the candidate splits).
Therefore, the extra overhead introduced by vertical partitioning is the sum of repartition time and label broadcasting time,
which is only 10\% of data loading and sketching on small dataset like \textit{RCV1} and 24\% on large dataset like \textit{Synthesis}.
The extra overhead in vertical partitioning is worth-while given the overall performance improvement.

\begin{table}[!h]
\tiny
\centering
\begin{tabular}{c c c c c c c}
\toprule[1.5pt]
\multirow{2}*{Dataset} & \multirow{2}*{\specialcell{Load\\Data}} & \multirow{2}*{\specialcell{Get\\Splits}} & \multicolumn{3}{c}{Repartition} & \multirow{2}*{\specialcell{Broadcast\\Label}} \\
\cmidrule[1pt](l{2pt}r{2pt}){4-6}
& & & Na\"ive & Compress & \alg & \\
\midrule[1pt]
RCV1 & 17 & 2 & 7 & 4 & 2 & 0.4 \\
\midrule[1pt]
RCV1-multi & 12 & 2 & 5 & 3 & 2 & 0.3 \\
\midrule[1pt]
Synthesis & 584 & 65 & 329 & 276 & 158 & 6 \\
\bottomrule[1.5pt]
\end{tabular}
\caption{\small Time cost (in seconds) for data loading and preprocessing. We run three times and report the average.}
\label{tb:transform_expr}
\vspace{-10pt}
\end{table}

\section{Scalability of \alg}
\label{sec:scalability}
We further conduct an experiment to assess the scalability of \alg. 
Since the \textit{Synthesis} dataset cannot fit in memory of two machines, we use two subsets of it, as Section~\ref{sec:quadrant_expr} does.
Specifically, \textit{Synthesis-N10M} refers to the subset of the first 10 million instances and 
\textit{Synthesis-D25K} the subset of the first 25 thousand features.
We present the results in Table~\ref{tb:scalability}.
In overall, \alg runs faster given more machines.
However, linear speedup is not observed on both datasets, since the time cost of some operations in \alg have no relations to number of machines.
For instance, in node splitting, all workers need to update the position of every instance, which is not able to speedup given more workers.
Therefore, the speedup on \textit{Synthesis-D25K} is lower as it contains more instances, while on \textit{Synthesis-N10M} we can achieve higher speedup.
However, we can accelerate such computation with multi-threading.
Since the memory consumption of \alg is much smaller than the horizontal-based implementations, we should consider 
using a small number of machines with multiple CPU cores
to achieve higher speedup. 

\begin{table}[!ht]
\scriptsize
\centering
\begin{tabular}{ccccccccc}
\toprule[1.5pt]
Dataset &
\multicolumn{4}{c}{Synthesis-N10M} & \multicolumn{4}{c}{Synthesis-D25K} \\
\midrule[1pt]
\# Machine & 2 & 4 & 6 & 8 & 2 & 4 & 6 & 8 \\
\midrule[1pt]
Run time & 32.2 & 18.6 & 13.7 & 12.5 & 32.1 & 25.7 & 23.4 & 20.2 \\
\midrule[1pt]
Speedup & 1.0 & 1.7 & 2.4 & 2.6 & 1.0 & 1.2 & 1.4 & 1.6 \\
\bottomrule[1.5pt]
\end{tabular}
\caption{\small Scalability test. Run time in seconds.}
\vspace{-10pt}
\label{tb:scalability}
\end{table}

\section{Comparison with Yggdrasil}
\label{sec:appendix_yggdrasil}

Since Yggdrasil can only train vanilla decision trees on low dimensional datasets, we implement a representative of QD3 in Section~\ref{sec:eval} and assess the impact of storage pattern.
To validate the ability of our implementation to represent QD3, we compare it with Yggdrasil in this section.

The experiments are carried out on three low dimensional datasets listed in Table~\ref{tb:low_dim_dataset}.
We use 5 workers for all three datasets and other hyper-parameters are the same as in Section~\ref{sec:eval}.
The results are also given in Table~\ref{tb:low_dim_dataset}.
As aforementioned, we combine instance-to-node index and node-to-instance index for optimization, 
therefore, our implementation in QD3 is able to outperform Yggdrasil on the three datasets.
In addition, \alg is the fastest, verifying the QD4 is more computation-efficient owing the row-store it adopts.

\begin{table}[h]
\scriptsize
\centering
\begin{tabular}{c c c c c}
\toprule[1.5pt]
Dataset & Size & Yggdrasil & QD3 (Ours) & \alg \\
\midrule[1pt]
Epsilon & N=500K D=2K & 137 & 24 & 5 \\
\midrule[1pt]
SUSY & N=5M D=18 & 32 & 9 & 5 \\
\midrule[1pt]
Higgs & N=11M D=28 & 71 & 14 & 7 \\
\bottomrule[1.5pt]
\end{tabular}
\caption{\small Experiments on low dimensional datasets. The rightmost three columns are time cost for one tree in seconds.}
\label{tb:low_dim_dataset}
\vspace{-10pt}
\end{table}

\section{Comparison with LightGBM}
\label{sec:appendix_lgb}

LightGBM supports both data-parallel and feature-parallel strategies.
Data-parallel horizontally partitions the dataset onto workers and stores the data in row-manner,
which is also chosen as our baseline in Section~\ref{sec:eval}.
Feature-parallel, however, does not partition the dataset.
It demands that all workers load a full copy of the dataset. In histogram construction and split finding, each worker independently builds histogram for a feature subset and finds the local best split, as vertical partitioning does.
In node splitting, each worker splits a node as the horizontal partitioning does, since it owns a full copy of dataset.
Although such approach can avoid heavy communication,
it only works for small-scale datasets.
For many real-world workloads, the size of dataset usually exceeds the memory of each machine, therefore the feature-parallel implementation of LightGBM is impractical.

Here we conduct experiments on two small datasets, \textit{RCV1} and \textit{RCV1-multi}.
As shown in Table~\ref{tb:small_dataset}, the feature-parallel version can outperform data-parallel, since it avoids the aggregation of histograms.
However, \alg still achieves the fastest speed.
Since the datasets contain smaller numbers of instances, the communication cost of \alg does not dominant the overall run time.
As a result, \alg is able to outperform the feature-parallel LightGBM on small-scale datasets.

\begin{table}[!h]
\scriptsize
\centering
\begin{tabular}{c c c c}
\toprule[1.5pt]
Dataset & LightGBM (DP) & LightGBM (FP) & \alg \\
\midrule[1pt]
RCV1 & 17 & 5 & 3 \\
\midrule[1pt]
RCV1-multi & 127 & 23 & 13 \\
\bottomrule[1.5pt]
\end{tabular}
\caption{\small Time cost per tree in seconds. DP and FP refer to data-parallel and feature-parallel, respectively.}
\label{tb:small_dataset}
\vspace{-10pt}
\end{table}

\fi

\end{document}